\documentclass{article} 
\PassOptionsToPackage{table}{xcolor}
\usepackage{iclr2026_conference,times}
\usepackage{amsmath,amssymb,amsthm}
\usepackage{multirow}
\usepackage[table]{xcolor}
\usepackage{graphicx}
\usepackage{booktabs}
\usepackage{hyperref}
\usepackage{url}
\usepackage{microtype}

\newtheorem{proposition}{Proposition}
\newtheorem{definition}{Definition}

\definecolor{collapse}{RGB}{180,0,0}

\definecolor{ok}{RGB}{0,120,50}
\definecolor{note}{RGB}{80,80,200}

\iclrfinalcopy

\title{Conformity Breaks Conformal Prediction}
\author{Yibo Hu \\ Illinois Institute of Technology \\ \texttt{yhu89@illinoistech.edu}
\And
Hanyu Su \\ Illinois Institute of Technology \\ \texttt{hsu18@hawk.illinoistech.edu}}

\begin{document}
\maketitle
\lhead{}

\begin{abstract}

A conformal certificate can be valid when an LLM answers alone and invalid when the
same LLM sees peers that unanimously assert a wrong answer. The question is unchanged;
the model's score for the correct answer changes. We call this a score-mechanism shift: clean calibration certifies how the model scores answers alone, but not how it scores them under peer pressure. We show that this shift silently breaks conformal prediction in multi-agent LLM systems. Across open-weight models and multiple-choice QA tasks, coverage falls from a calibrated 90\% to 74\% under unanimous-wrong peers at the standard $\alpha=0.10$ operating point. The average hides a sharper failure: by targeting the low-confidence items the certificate still covers, an attacker nearly halves coverage on that subgroup, from 87\% to 47\%, while the monitored average remains much higher. The failure also reaches the decision layer: a system that should escalate when uncertain can instead become confident enough to act on the attacker's wrong answer. Standard conformal fixes do not solve the problem, because the question distribution has not changed; the model's scoring behavior has.\footnote{Code and data: \url{https://github.com/yibo-hu-lab/conformity-breaks-conformal}}
\end{abstract}

\section{Introduction}
\label{sec:intro}

Conformal prediction gives any model a formal coverage guarantee: with
probability $1-\alpha$ the correct answer lies in a prediction set
\citep{vovk2005,angelopoulos2021}. The guarantee is distribution-free, but it rests on
one assumption: the calibration data and the test data are exchangeable. Calibrate on
one distribution, and the certificate is valid only on that distribution.

We show that multi-agent LLM systems break this assumption from the inside. A model is
calibrated answering alone, then deployed reading other agents' answers. When peers
unanimously assert a wrong answer, the model often abandons its own correct one: the
question is identical, and only its score for the correct answer moves. We call this a
\emph{score-mechanism shift}, and it is not the covariate shift the conformal toolkit
knows how to repair, because the input never changes.

An attacker who controls the peer channel can trigger it on demand, silently, without
touching the input. This matters most for the closest prior work, which moves in the
opposite direction: \citet{wang2026debate2decision} use conformal prediction to
defend multi-agent debate, with an act-vs-escalate policy that intercepts $81.9\%$
of wrong-consensus cases. But their guarantee is marginal, they treat peer influence
as benign, and they explicitly defer the conditional case. We show the deferred case
is where the attack lives.

\paragraph{Contributions.}
We study one mechanism and its consequences for clean-calibrated conformal prediction
in multi-agent LLM systems.
\begin{enumerate}
  \item \textbf{A new endogenous shift.} We identify social conformity as an
        adversarially controllable shift that voids clean-calibrated conformal coverage
        for LLMs, and name it the \emph{score-mechanism shift}
        (Definition~\ref{def:smshift}). Marginal coverage falls from $90\%$ to $74\%$
        under unanimous-wrong peers.
  \item \textbf{A hidden conditional failure.} On the low-confidence items the
        certificate still covers, coverage falls from $87\%$ to $47\%$ while the monitored
        average stays at a reassuring $76\%$. Without pressure that subgroup sits at $87\%$,
        so the loss is the attack's, not a calibration artifact. An adversary without the
        answer key takes $40$\,pp from it by attacking a chosen $10\%$ of items, moving
        the overall average by $4$ (Section~\ref{sec:e2}).
  \item \textbf{A failed act-vs-escalate defense.} Under the same shift, Wang et al.'s
        clean-calibrated act-vs-escalate layer commits the attacker's answer on $12\%$
        of flipped items for the pooled predictor, and on $71\%$ for the largest single
        model we test, Qwen2.5-32B (Section~\ref{sec:e1}).
\end{enumerate}

\section{Setup: Conformal Prediction under Peer Pressure}
\label{sec:setup}

\paragraph{Intuition.}
A single item shows the failure mode. Calibration on clean data fixes a rule like
``keep every answer the model gives probability at least $0.42$.'' Answering alone, the
model puts $0.55$ on the correct option, so it is covered. Show it peers that
confidently assert a wrong option, and its correct-answer probability drops to
$0.25$ while the wrong option rises to $0.70$: the same question now yields a set
that contains only the wrong answer. The question never changed; only the score did.
(Numbers illustrative.)

\paragraph{Conformal prediction.}
Take any LLM used as a multiple-choice QA system. We score each candidate answer by
its \emph{nonconformity}, one minus the probability the model assigns it, and keep a
\emph{prediction set} of every answer whose score is at most a calibrated threshold
$\hat\tau$, that is $C(x)=\{y:p[y]\ge 1-\hat\tau\}$. The threshold $\hat\tau$ is the
$(1-\alpha)$-quantile, on a held-out calibration set, of the scores of the
\emph{correct} answers, chosen so that the correct answer is included at least a
$(1-\alpha)$ fraction of the time (e.g.\ $90\%$). This is split (inductive) conformal
prediction with the LAC (least-ambiguous-classifier) score
\citep{vovk2005,angelopoulos2021}; it applies to any model with no distributional
assumptions, as long as calibration and test data are exchangeable. We use LAC
throughout. We also report the alternative APS (adaptive prediction sets) rule, which
degenerates on these four-option distributions and is discussed where it appears
(Sections~\ref{sec:e2} and~\ref{sec:e1}).

\paragraph{What the input $x$ is.}
Throughout, $x$ is the question, the task instance the certificate is about, and the
score is the model's probability over its answer options. The peer transcript is not
part of $x$; it is part of the context that generates the score. This is the natural
choice for the deployment we study: the guarantee is sold as a statement about
questions, the defender calibrates on questions, and the peer channel is exogenous, not
a feature the certificate ranges over. The question is therefore identical at
calibration and test, so the shift is in the score mechanism, not in $x$; we revisit
the alternative of folding the transcript into $x$ in Section~\ref{sec:tibshirani}.

\paragraph{Social conformity.}
In multi-agent setups, LLMs routinely see other LLMs' answers. We and others
\citep{qu2026easiermislead} have shown that when several ``peer'' LLMs all give
the same wrong answer, the target LLM often abandons its own correct answer to
match them. We call the fraction of items where this behavioral switch happens the
\emph{answer-flip rate}; for Qwen2.5-7B under unanimous-wrong peers it is about
$75\%$. The answer-flip rate is a behavioral quantity and is distinct from the
\emph{coverage-loss rate}, the fraction of items whose score crosses $\hat\tau$;
the two closely track each other under the unanimous-wrong condition we study, for
reasons we make precise in Section~\ref{sec:prop}.

The certificate is fixed on solo data but deployed under peer pressure, so when
pressure pushes probability mass off the correct answer, items once inside the
prediction set fall out and the certificate breaks with no visible signal.

\paragraph{Data.}
For each (model, question) pair we record the model's solo (round-1) answer
distribution and its distribution under six peer-pressure conditions: unanimous wrong,
mixed, and unanimous correct, each with and without an added authority cue. In each
condition the model re-sees the question alongside a fabricated conversation in which
``peer'' LLMs all give the same answer, six turns of the form
``Mary: I think this answer is (B).''
We use four core open models
(Qwen2.5-7B, Llama-3.1-8B, Mistral-7B-v0.3, Gemma-2-9B) on ARC-Challenge and
TruthfulQA, with two further models and five more task types in two of three
seeds. Full model lists, dataset counts, and hardware are in
Appendix~\ref{app:robgen}.

\paragraph{Conformal procedure.}
Calibration uses round-1 (solo) scores $s_i = 1 - p_{\mathrm{gt}}^{(\mathrm{r1})}$; a
test item under condition $c$ is covered iff $p_{\mathrm{gt}}^{(c)} \ge 1-\hat\tau$,
and the coverage rate is the fraction of covered test items. We average over $2{,}000$
random $50/50$ calibration/test splits stratified by (model, dataset).

\paragraph{Headline condition.}
Our headline numbers use the \emph{no-authority} variant of the unanimous-wrong
condition: the peer transcript asserts the wrong answer with no appeal to
authority. The $+$authority variant behaves almost identically (pooled $74.6\%$
vs.\ $73.9\%$ at $\alpha=0.10$) and is reported alongside in
Table~\ref{tab:main}.\footnote{The conditional and act-vs-escalate results
(Sections~\ref{sec:e2}--\ref{sec:e1}) use the authority cue; repeating them without
it gives the same conditional collapse and a slightly stronger gate-defeat, so the
cue does not drive the effect.}

\section{Main Results}
\label{sec:results}

We report three failures of increasing severity: a marginal coverage collapse, then a
hidden conditional collapse on the attacker-targeted subgroup, then the failure of a
clean-calibrated act-vs-escalate policy. Two units of analysis appear below.
Section~\ref{sec:marginal} scores each (model, question) pair against a single threshold
shared by the four models. Sections~\ref{sec:e2} and~\ref{sec:e1} score the four models
as one \emph{pooled predictor}: a uniform-weight opinion pool of their per-option
distributions, calibrated as a single system, which is the object a deployer certifies.

\subsection{Unanimous-wrong peers break marginal coverage}
\label{sec:marginal}

Calibrating on round-1 (no-pressure) data and testing under each condition, the
certificate holds everywhere except unanimous wrong (Table~\ref{tab:main}). At the
standard $\alpha=0.10$ operating point, coverage under unanimous-wrong pressure
falls to \textcolor{collapse}{$73.9\%\pm0.3\%$} (mean over three seeds), a stable
$-16.1$\,pp gap, while the round-1 certificate stays right at target
($90.0\%\pm0.0\%$).

\begin{table}[h]
\centering
\caption{Coverage collapses only under unanimous-wrong peers. Three seeds
(common subset: 4 models $\times$ ARC $+$ TruthfulQA), target $90\%$
($\alpha=0.10$) or $95\%$ ($\alpha=0.05$).}
\label{tab:main}
\small
\begin{tabular}{lcccccc}
\toprule
\multirow{2}{*}{Condition} &
\multicolumn{3}{c}{$\alpha = 0.10$ (target 90\%)} &
\multicolumn{3}{c}{$\alpha = 0.05$ (target 95\%)} \\
\cmidrule(lr){2-4}\cmidrule(lr){5-7}
& seed0 & seed1 & seed2 & seed0 & seed1 & seed2 \\
\midrule
Round 1 (no pressure) & 90.1\% & 90.0\% & 90.0\% & 95.0\% & 95.0\% & 95.0\% \\
\rowcolor{red!8}
Unan.\ wrong (no auth.) &
  \textcolor{collapse}{73.5\%} &
  \textcolor{collapse}{74.1\%} &
  \textcolor{collapse}{74.2\%} &
  \textcolor{collapse}{85.3\%} &
  \textcolor{collapse}{86.6\%} &
  \textcolor{collapse}{87.1\%} \\
\rowcolor{red!8}
Unan.\ wrong (+ auth.) &
  \textcolor{collapse}{75.1\%} &
  \textcolor{collapse}{74.0\%} &
  \textcolor{collapse}{74.6\%} &
  \textcolor{collapse}{87.1\%} &
  \textcolor{collapse}{87.4\%} &
  \textcolor{collapse}{86.7\%} \\
Mixed (no auth.) & 94.2\% & 94.0\% & 94.4\% & 98.4\% & 99.0\% & 99.2\% \\
Mixed (+ auth.) & 93.4\% & 93.5\% & 94.1\% & 98.6\% & 98.9\% & 98.9\% \\
Unan.\ correct (no auth.) & 93.5\% & 94.5\% & 94.3\% & 99.5\% & 99.9\% & 100.0\% \\
Unan.\ correct (+ auth.) & 93.5\% & 94.5\% & 94.4\% & 99.5\% & 100.0\% & 99.9\% \\
\bottomrule
\end{tabular}
\end{table}

\begin{figure}[t]
  \centering
  \includegraphics[width=0.62\textwidth]{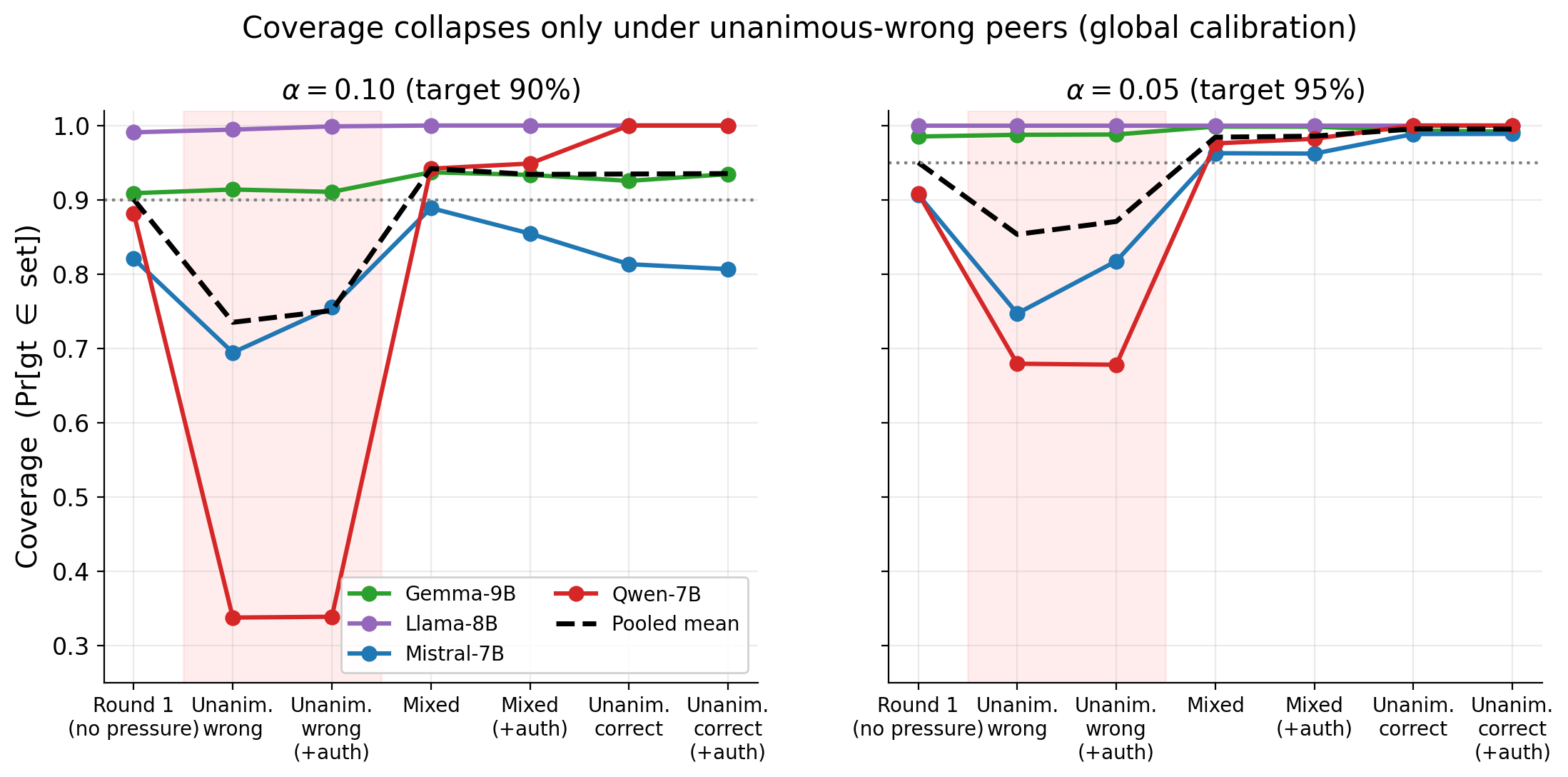}
  \caption{Coverage collapses only under unanimous-wrong peers and improves under
    unanimous-correct ones. Per model; $\alpha=0.10$ (left), $\alpha=0.05$ (right).
    Llama's immunity is a global-calibration artifact (Section~\ref{sec:mondrian}).}
  \label{fig:gap}
\end{figure}

The collapse depends on the target level, and is largest where practitioners
deploy: $-16.1$\,pp at $\alpha=0.10$, $-8.6$\,pp at $\alpha=0.05$, and vanishing at
$\alpha=0.01$, where the threshold is so loose the set is nearly vacuous. Unanimous
wrong is the sole condition that collapses coverage; unanimous correct \emph{improves}
it (Figure~\ref{fig:gap}).

The marginal collapse is the easiest result to state, but on its own it understates
the danger and sits adjacent to prior work that uses conformal prediction to
\emph{defend} this setting (\citealp{wang2026debate2decision}; Section~\ref{sec:related}). The next two results
are the paper's core contribution: a conditional collapse hidden behind a healthy
marginal number, and a direct defeat of the published act-vs-escalate defense, both at
$\alpha=0.10$.

\subsection{Targeted pressure creates a hidden conditional collapse}
\label{sec:e2}

Marginal coverage is the number a deployer monitors. We instead ask how coverage
behaves on the subgroup an attacker would target. Coverage can only be lost where it was
held, so the subgroup of interest is the low-confidence items the certificate still
covers. At $\alpha=0.10$ the threshold falls near the tenth percentile of solo confidence
(round-1 $p_{\mathrm{gt}}$): below it the certificate covers only $13\%$ of items even
without pressure, while above it coverage is near-complete. We therefore report the
decile immediately above the threshold, which we call the \emph{targeted band}; at
$\alpha=0.05$ the corresponding band is the fifth to tenth percentile. The predictor is
split-conformal LAC, calibrated on clean round-1 data.

For the pooled four-model predictor, conformity injection leaves marginal coverage only
mildly degraded ($90.1\%\!\to\!75.7\%$), so the monitored number stays a reassuring
$76\%$. But on the targeted band coverage falls from \textbf{87.3\%} to
\textbf{47.4\%} (Figure~\ref{fig:e2}; per-model numbers in
Appendix Table~\ref{tab:conditional}). The collapse lands exactly where the certificate is
least useful: the low-confidence items a user most needs it for. Without pressure the band
sits within three points of the $90\%$ target. The loss is therefore the attack's, not the
distribution-free conditional-coverage limit \citet{barber2021} formalize. The new
element is that the failing subgroup is attacker-targetable rather than incidental, the
conditional case \citet{wang2026debate2decision} defer.

\begin{figure}[h]
  \centering
  \includegraphics[width=0.95\textwidth]{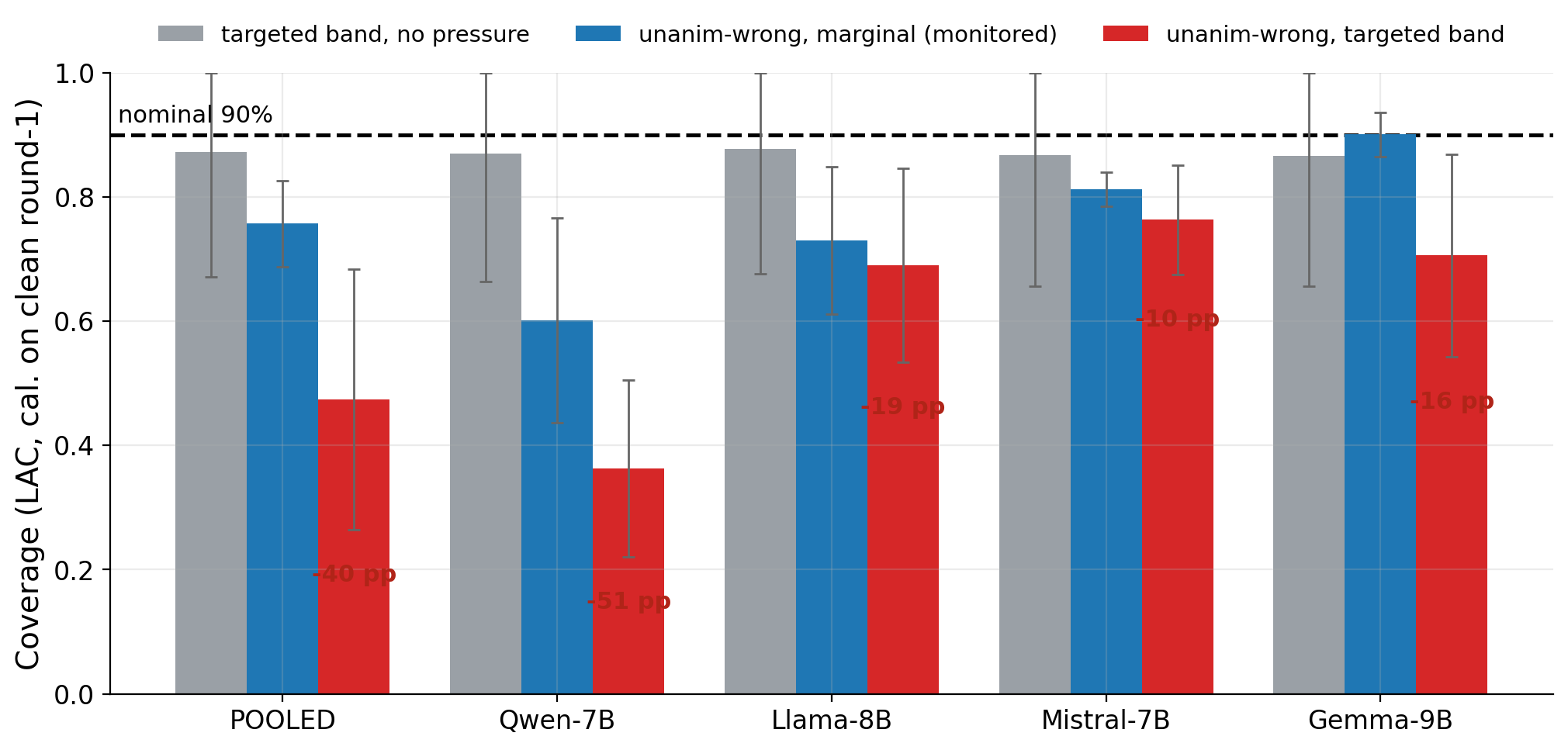}
  \caption{Conditional coverage collapses while marginal coverage holds. The targeted
    band is the decile of solo confidence immediately above the calibrated threshold, where
    the certificate covers $87\%$ of items without pressure. Under unanimous-wrong pressure
    the pooled predictor keeps marginal coverage at $76\%$ while coverage on the band falls
    to $47\%$; every model drops. Gemma-9B is the sharpest illustration: its monitored
    marginal coverage is exactly at target, and its band is at $71\%$. Error bars are
    standard deviations over $2{,}000$ calibration splits.}
  \label{fig:e2}
\end{figure}

\paragraph{Targeting it needs no privileged information.}
The attacker we consider (Section~\ref{sec:threat}) chooses which items to attack, and
an intelligent one attacks the items the model is least sure of. What ``least sure'' means decides
whether the attack works. Ranking by $p_{\mathrm{gt}}$ needs the answer key, and its
bottom decile is the decile the threshold already excludes: coverage there is $13.4\%$
before any attack, so attacking it gains nothing. Ranking by what the deployment actually
reports needs no answer key and is far more damaging.

Three adversaries, none of whom knows the correct answers, all reach the items conformal
still covers. One probes the deployment with clean queries and attacks the
least-confident $10\%$ by that signal, a set that overlaps the band above without
coinciding with it: coverage there falls from $92.2\%$ to $51.9\%$
($\mathbf{-40.2}$\,pp). One sees only the prediction sets, not the scores, and attacks
the largest ones: $-37.9$\,pp. One never touches the deployment at all and ranks with an
outside model of comparable quality (Qwen-32B, not part of the ensemble): $-36.4$\,pp.
Attacking a random $10\%$ costs $-14.3$\,pp (Figure~\ref{fig:adversary}). Attacking only
$10\%$ of items moves overall coverage from $90.1\%$ to $86.0\%$: the adversary buys a
large failure on a small chosen subset while the monitored average stays within a few
points of target.

The band collapse of Figure~\ref{fig:e2} holds for every model individually, from
$-10$\,pp for Mistral-7B to $-51$\,pp for Qwen-7B. It does not appear under APS, where
coverage on the corresponding band moves by $-3$ to $+6$\,pp, but that reflects a
degenerate set rule rather than a robust one: APS sets in this data average $4.0$ of $4$
options, so nothing can fall out of them. Section~\ref{sec:e1} quantifies the same
degeneracy at the decision layer. Our results are therefore statements about LAC, the
standard and least conservative of the two rules. The pooled
predictor falls further than three of the four models it averages: the four see the same
fabricated transcript, so their shifts are common-mode and pooling does not cancel them.
The wider bottom-$20\%$ band straddles the threshold and mixes this loss with a floor effect on
items the certificate never covered; it understates the attack ($50.3\%$ to $35.5\%$).
Per-model numbers, the bottom-$k\%$ bands, and a breakdown by confidence stratum are in
Appendix~\ref{app:condfull}.

\subsection{The clean-calibrated act-vs-escalate defense fails}
\label{sec:e1}

We reproduce the relevant defense mechanism from \citet{wang2026debate2decision}: a
uniform-weight linear opinion pool of the four models' full per-option
distributions, split-conformal LAC sets calibrated on the clean pooled round-1
distribution, and their act-vs-escalate policy, which \emph{acts} (commits the answer
autonomously) when the prediction set is a singleton and \emph{escalates} to a
human when the set has size $\ge 2$. Injecting unanimous-wrong conformity, we
measure how often the set collapses to a singleton \emph{equal to the wrong answer},
among items where conformity flips the pooled top-1 from correct to wrong. On those
items the safety layer commits the attacker's answer and bypasses escalation.

For the pooled predictor, \textbf{12.1\%} of conformity-flipped items
(seed-0, authority cue; a median of $64$ flipped items per split, $\pm5.5$ over
$2{,}000$ calibration splits) yield a singleton set equal to the wrong answer, so the
clean-calibrated safety layer acts on the attacker's answer with no escalation; per
single model the rate reaches $28.7\%$ (Qwen) and $26.9\%$ (Mistral), while Gemma is
near zero because its sets rarely collapse: on flipped items they average $3.0$ of $4$
options, against $2.4$ for Qwen (Table~\ref{tab:e1}). The defeat replicates
across seeds and is not authority-driven (pooled $13.5\%\pm1.0$ with
the authority cue, $15.8\%\pm1.2$ without; these spreads are over seeds, not splits).

Under APS the rate falls to zero, but the APS gate is inoperative here rather than
defended: its calibrated quantile reaches $0.9997$, sets average $4.0$ of $4$ options,
and the layer acts autonomously on $4.5\%$ of items even without pressure, against
$44.6\%$ under LAC.

The largest open model we test, Qwen2.5-32B (seed-0, ARC and TruthfulQA), sharpens
the pattern. On its own threshold its marginal coverage under pressure is $63.0\%$,
below every model in Table~\ref{tab:conditional} except Qwen-7B; coverage on its
targeted band
falls from $87.2\%$ to $30.3\%$; and its singleton-act-on-wrong rate reaches
\textcolor{collapse}{$71.4\%$}, two and a half times the highest entry in
Table~\ref{tab:e1}.
It is also the most confident model in the set before any pressure is applied (mean
$p_{\mathrm{gt}}^{(r1)}=0.883$, against $0.743$ for the robust Gemma-9B). Neither scale
nor confidence buys protection, and the most capable model is the most dangerous at the
action layer.

\begin{table}[h]
\centering
\caption{The act-vs-escalate defense fails. Among items where conformity flips the
pooled top-1 from correct to wrong, the fraction yielding a singleton set equal to
the \emph{wrong} answer, so the clean-calibrated layer commits the attacker's answer
with no escalation (the complement either escalates or acts on the correct answer).
LAC, $\alpha=0.10$, seed-0, with the authority cue.}
\label{tab:e1}
\small
\begin{tabular}{lc}
\toprule
Unit & Singleton-ACT on wrong answer (\%) \\
\midrule
\rowcolor{red!8}
\textbf{POOLED} & \textcolor{collapse}{\textbf{12.1}} \\
Qwen-7B & \textcolor{collapse}{28.7} \\
Mistral-7B & \textcolor{collapse}{26.9} \\
Llama-8B & \textcolor{collapse}{17.5} \\
Gemma-9B & \textcolor{ok}{0.2} \\
\bottomrule
\end{tabular}
\end{table}

Tightening $\alpha$ helps but does not close the failure. At $\alpha=0.05$ the
singleton-ACT rate falls from $12.1\%$ to $1.9\%$ and the targeted band, now the decile
above the tighter threshold, sits $21.3$\,pp under its $95\%$ target instead of the
$42.6$\,pp under $90\%$ at $\alpha=0.10$: half the hole, still a hole. It is also not
free, since the average pressured set grows from $2.44$ to $3.31$ of four options and
the layer acts autonomously on $2.6\%$ of pressured items rather than $9.2\%$. No single
offline-calibrated $\alpha$ closes both (Section~\ref{sec:threat}).

\section{Why the Failure Occurs and Why Standard Fixes Do Not Suffice}
\label{sec:why}

The collapses of Section~\ref{sec:results} have a single cause: peer pressure
changes the model's score distribution at a fixed input. We first name this shift
and quantify the coverage loss it produces (Section~\ref{sec:prop}), then show that
the two textbook conformal repairs do not touch it (Sections~\ref{sec:tibshirani},
\ref{sec:mondrian}), and finally that the one repair that restores coverage is not a
standard offline fix (Section~\ref{sec:condaware}).

\subsection{The shift is in the score mechanism, not the input}
\label{sec:prop}

\begin{definition}[Score-mechanism shift]
\label{def:smshift}
Let $s(x,y)$ be the nonconformity score and let the marginal distribution of the
input $x$ be \emph{fixed} between calibration and test. A \emph{score-mechanism
shift} is a change in the conditional law of the score $s$ given $x$, equivalently
a change in $p_{\mathrm{gt}}(x,\cdot)$ for fixed $x$, caused by a change in the
surrounding \emph{context} rather than in the input. Here the context is social: the
model is calibrated answering alone and tested under peer pressure. The question is
identical, so the input distribution is unchanged and only the model's score moves.
\end{definition}

We now quantify the coverage loss such a shift produces. The formal point is simple:
if peer pressure pushes the nonconformity scores upward, the coverage loss is exactly
the \emph{net} probability mass that crosses the clean calibrated threshold. Let $\mathcal{D}_0$
be the distribution of scores $s = 1 - p_{\mathrm{gt}}$ when the model answers alone, with
CDF $F_0$, and let $\hat\tau = F_0^{-1}(1-\alpha)$ be the calibrated threshold. Let
$\mathcal{D}_1$, with CDF $F_1$, be the score distribution under unanimous-wrong
pressure. Coverage under pressure is $\Pr_{s\sim\mathcal{D}_1}[s\le\hat\tau]
= F_1(\hat\tau)$.

\begin{proposition}[Coverage loss under a score-mechanism shift]
\label{prop:coverage}
\textup{(i)} Coupling the clean and pressured scores on each item, and with no
assumption on the direction of the shift, the coverage gap equals the \emph{net}
threshold-crossing flow:
\[
  \Delta = F_0(\hat\tau) - F_1(\hat\tau)
  = \underbrace{\Pr[s_0\le\hat\tau<s_1]}_{\text{down-cross (covered}\to\text{uncovered)}}
  - \underbrace{\Pr[s_1\le\hat\tau<s_0]}_{\text{up-cross (uncovered}\to\text{covered})}
  =: \mathrm{CoverageLossRate}(\hat\tau).
\]
\textup{(ii)} If in addition peer pressure raises the score distribution at the
calibrated threshold, i.e.\ $F_1(\hat\tau)\le F_0(\hat\tau)$, then
$\mathrm{Coverage}(\hat\tau,\mathcal{D}_1) = F_1(\hat\tau) \le 1-\alpha$.
First-order stochastic dominance of $\mathcal{D}_1$ over $\mathcal{D}_0$ is sufficient
but not necessary: only the value at $\hat\tau$ enters.
\end{proposition}

Part (i) follows by counting threshold crossings item by item; part (ii) is immediate
from the definition of $\hat\tau$ (proof in Appendix~\ref{app:proof}).

The dominance hypothesis is an empirical fact here, not a modelling convenience: under
unanimous-wrong peers $F_1(t)\le F_0(t)$ holds at every $t$, while under mixed and
unanimous-correct peers it fails at $17\%$ and $28\%$ of thresholds respectively. The
decomposition, which needs no such hypothesis, applies throughout.

Table~\ref{tab:flow} counts the flow. Under unanimous-wrong peers $19.8\%$ of covered
items are pushed out and $3.2\%$ are pulled in, so the net loss is $16.6$\,pp, matching
the measured coverage gap exactly. The same accounting explains the conditions in which
coverage \emph{rises} above nominal: under mixed and unanimous-correct peers the inflow
exceeds the outflow, so the gap is negative. Up-crossings are a minority throughout, but
not a rounding error.

\begin{table}[h]
\centering
\caption{Threshold-crossing flow, scored per (model, question) pair against the shared
threshold of Section~\ref{sec:marginal} ($\alpha=0.10$, seed-0, mean over $2{,}000$
splits). Net flow equals the coverage gap in every condition, as
Proposition~\ref{prop:coverage}(i) requires.}
\label{tab:flow}
\small
\begin{tabular}{lcccc}
\toprule
Condition & Down-cross & Up-cross & Net & Coverage gap \\
\midrule
\rowcolor{red!8}
Unan.\ wrong & 19.8\% & 3.2\% & \textcolor{collapse}{$+$16.6\,pp} & \textcolor{collapse}{$+$16.6\,pp} \\
\rowcolor{red!8}
Unan.\ wrong ($+$auth.) & 19.4\% & 4.5\% & \textcolor{collapse}{$+$15.0\,pp} & \textcolor{collapse}{$+$15.0\,pp} \\
Mixed & 2.0\% & 6.1\% & $-$4.1\,pp & $-$4.1\,pp \\
Mixed ($+$auth.) & 2.7\% & 6.0\% & $-$3.4\,pp & $-$3.4\,pp \\
Unan.\ correct & 3.9\% & 7.3\% & $-$3.4\,pp & $-$3.4\,pp \\
Unan.\ correct ($+$auth.) & 3.6\% & 7.1\% & $-$3.5\,pp & $-$3.5\,pp \\
\bottomrule
\end{tabular}
\end{table}

The coverage-loss rate is threshold-relative and so
distinct from the behavioral answer-flip rate of Section~\ref{sec:setup}, but the two
closely track each other here (Figure~\ref{fig:robustness}, Panel D), because a flip is
what drives $p_{\mathrm{gt}}$ below the calibration mass; this is why our high-flip
models are our high-coverage-loss models.

\subsection{Weighted conformal prediction targets the wrong shift}
\label{sec:tibshirani}

The first textbook repair is weighted conformal prediction. \citet{tibshirani2019}
recover valid coverage under covariate shift by reweighting each calibration point
by the likelihood ratio $w(x)=p_{\mathrm{test}}(x)/p_{\mathrm{cal}}(x)$. This does
not help here, because our shift is not a covariate shift: the question $x$ is
identical at calibration and test, so $w(x)=1$ for every $x$ and the correction
reduces to the identity. What changes is the model's own output distribution given
$x$, a score-mechanism shift (Definition~\ref{def:smshift}), not the covariate
marginal the weight corrects. Attack status is not a function of $x$, so no weight
function can separate pressured from clean items: even an oracle that knew in advance
which items would face pressure would have to assign them the same weight. Weighted
conformal is the identity here regardless of what the defender knows, and vanilla,
$w(x)=1$, and oracle weighting return identical coverage in our data.

This conclusion takes $x$ to be the question (Section~\ref{sec:setup}). Folding the
peer transcript into $x$ makes the same effect a covariate shift in the enlarged space,
where weighted CP would apply given the likelihood ratio between pressured and clean
contexts. That ratio has no referent here. The attacker chooses the transcript and can
change it after seeing any estimate the defender forms, so there is no fixed
pressured-context distribution to take a ratio against. The obstacle is not estimation
but the absence of a fixed test distribution: this is a worst-case problem, not a
reweighting one.

\subsection{The shift survives per-model calibration}
\label{sec:mondrian}

The second textbook repair is to calibrate each model separately. Mondrian
conformal prediction \citep{mondrian} does exactly this, computing a separate
threshold $\hat\tau_m$ for each model $m$ from that model's own round-1 data. At the
standard $\alpha=0.10$ operating point, Mondrian removes one artifact but not the
collapse. It dissolves Llama's apparent immunity, though not for the obvious reason.
The shared threshold is set by the tenth percentile of the pooled clean scores, not by
their mean, and it is the \emph{other} three models that push that percentile down:
$12$--$21\%$ of their items carry $p_{\mathrm{gt}}<0.001$, against $1.7\%$ of Llama's.
Llama is a beneficiary of the loose threshold, not its cause. It in fact tightens the
pool: removing it loosens the shared threshold further, from $0.000388$ to $0.000098$,
and its own threshold is the strictest of the four (Table~\ref{tab:tau}).

Under that stricter threshold Llama drops to $\approx72\%$. Qwen improves from $34\%$
to $61\%$ but still collapses, and Mistral lands at $\approx74\%$, so three of four
models still fail (Figure~\ref{fig:mondrian}, Appendix~\ref{app:robgen}).

\begin{table}[h]
\centering
\caption{Calibrated thresholds on the probability scale (an answer is kept when
$p\ge$ the entry; $\alpha=0.10$, seed-0, mean over $2{,}000$ splits). Low thresholds
come from items the model scores near zero, not from low average confidence.}
\label{tab:tau}
\small
\begin{tabular}{lccc}
\toprule
Unit & Threshold & Mean $p_{\mathrm{gt}}^{(r1)}$ & Items with $p_{\mathrm{gt}}^{(r1)}<0.001$ \\
\midrule
Shared (four models) & 0.000388 & --- & --- \\
\midrule
\rowcolor{red!8}
Llama-8B & 0.012596 & 0.557 & \textcolor{ok}{1.7\%} \\
Gemma-9B & 0.000499 & 0.743 & 12.3\% \\
Qwen-7B & 0.000124 & 0.753 & 13.7\% \\
Mistral-7B & 0.000025 & 0.634 & \textcolor{collapse}{20.7\%} \\
\bottomrule
\end{tabular}
\end{table}

The $\alpha$-dependence is worth stating precisely. At
the tighter $\alpha=0.05$ level, per-model calibration largely rescues the two
vulnerable models (Mondrian lifts Qwen to $\approx90\%$ and Mistral to $\approx
88\%$); at that operating point the collapse is mostly a pooled-calibration
artifact. It is at the standard $\alpha=0.10$ level that the collapse survives fair
per-model calibration, which is why it reflects genuine conformity behavior rather than
an artifact of pooling heterogeneous models.

\subsection{Restoring coverage requires the pressure labels the attack denies}
\label{sec:condaware}

The one repair that does restore coverage is to calibrate the predictor on
under-pressure data (condition-aware Mondrian). It restores exchangeability by
construction, and marginal coverage returns to $90.0\%$; even the targeted band of
Section~\ref{sec:e2} recovers, from $47.4\%$ to $83.9\%$. It counts as a standard
offline fix only when the defender can anticipate or label the pressured condition,
that is, knows in advance which items the adversary will attack. If the defender could
label those items, the attack would already be detected and those items could be
dropped. We make this argument precise in the threat model
(Section~\ref{sec:threat}).

The repair works by widening the sets, not by resisting the attack.
Calibrating on pressured scores loosens the threshold from $0.103$ to $0.009$, and the
average set grows from $2.4$ to $3.1$ of $4$ options. Under the act-vs-escalate rule of
Section~\ref{sec:e1} the layer then acts autonomously on $3.6\%$ of items, against
$44.6\%$ with no pressure and clean calibration: coverage is bought by escalating
almost everything. The one effective fix is effective because it gives up the autonomy
the layer exists to provide.

\section{Robustness and Generalization}
\label{sec:robustness}

The collapse is a genuine conformity effect rather than an artifact of how we measure or
calibrate it. It survives fair per-model calibration (per-model Mondrian still collapses
three of four models at $\alpha=0.10$, Section~\ref{sec:mondrian}); it is invariant to the
covariate-shift correction ($w(x)=1$ when $x$ is the question, so weighted CP reduces to
the identity, Section~\ref{sec:tibshirani}); it reflects the clean-versus-pressured
calibration mismatch rather than the choice of metric (calibrating and deploying on the
pressured distribution itself restores $90.0\%$ marginal coverage and cuts the
act-on-wrong rate from $12.1\%$ to $2.6\%$, Section~\ref{sec:condaware}); and it is
driven by realized score movement
rather than answer flips (the gap is measured on realized scores, and Qwen collapses
despite a larger calibration buffer than the robust Gemma, Section~\ref{sec:hetero}).
The subsections below ask which models are vulnerable, why, and whether that can be
measured in advance (Section~\ref{sec:hetero}), and show that the collapse generalizes
across task types (Section~\ref{sec:perdataset}).

\subsection{Vulnerability tracks score movement, and a stress test can rank it}
\label{sec:hetero}

Vulnerability is strongly model-dependent. We read it off per-model thresholds rather
than the shared one, so that the global-calibration artifact of
Section~\ref{sec:mondrian} does not contaminate the comparison: under its own threshold
Qwen-7B falls to $61\%$, Llama-8B to $72\%$, Mistral-7B to $74\%$, and Gemma-9B holds at
$90\%$. Under the shared threshold Llama instead reads $99.3\%$, which is the artifact,
not a property of the model.\footnote{Per-model flip rates are quoted as three-seed pooled values
(Gemma $16.2\%$, Qwen $75.8\%$; Table~\ref{tab:robustness}). The Qwen $53.7\%$ in
Appendix~\ref{app:qwensize} is a different data regime (seeds~1--2, all seven task
types) and is labeled there.}

The takeaway is that the driver is not how confident a model is but how far peer
pressure moves its correct-answer probability. Gemma and Qwen start from nearly
identical solo confidence, yet under pressure Qwen's $p_{\mathrm{gt}}$ drops by $0.747$
on average and Gemma's by only $0.151$ (Table~\ref{tab:robustness}, decomposed in
Figure~\ref{fig:robustness}). Qwen2.5-32B makes the same point from the other end: it
is the most confident model we test before any peers appear, and the second least
covered after they arrive (Section~\ref{sec:e1}).

This dynamic quantity, $\Delta p_{\mathrm{gt}}$, is measurable before deployment, and a
stress test built on it ranks models by vulnerability before any of them is calibrated.
Static properties do not: over the six models of each data regime, mean solo confidence and clean accuracy
rank pressured coverage at $\rho=-0.20$ and $+0.37$, the sign not even stable, while the
answer-flip rate ranks it at $-0.89$ and $-0.71$. The ranking also transfers across
tasks. Measuring the flip rate on one task type and predicting pressured coverage on a
different one gives a median $\rho=-0.49$ over the $42$ ordered pairs, $35$ of them in
the right direction (sign test $p<10^{-5}$), which is what we also get when both are
read off the same items: the association is not an artifact of measuring diagnostic and
outcome on one set of questions. Geometric shapes is the exception, the one task whose
pressured coverage no source task predicts. Of the two stress-test quantities the flip
rate ranks more reliably; $\Delta p_{\mathrm{gt}}$ carries the mechanism but is less
consistent across regimes ($-0.89$ and $-0.49$). Appendix~\ref{app:screen} gives both
tables.

Marginal and subgroup robustness rank models differently: Mistral holds its band at
$76.3\%$ while Gemma, whose marginal coverage is exactly at target, falls to $70.6\%$
(Appendix~\ref{app:condfull}).

\subsection{The collapse grows across seven task types}
\label{sec:perdataset}

The collapse is not specific to ARC and TruthfulQA. On seeds~1 and~2 (6 models
$\times$ 7 task types) it is if anything \emph{larger}: coverage under
unanimous-wrong falls to $69.8\%$ (a $-20.2$\,pp gap), and every task type degrades
by $-5$ to $-28$\,pp (Appendix~\ref{app:robgen}, Table~\ref{tab:perdataset}).

\section{Threat Model and Deployment Implications}
\label{sec:threat}

The attacker controls the orchestration layer of a multi-agent LLM system, or
compromises one agent within it, and can craft the ``peer responses'' the target
model sees before it produces its final answer. The attacker has no access to the
target model's weights, logits, or calibration data; the attack is purely social.
Concretely, the attacker can (i) inject a fabricated transcript in which $k\ge1$
peer agents all assert the wrong answer, exactly the setup of our experiments; (ii)
select which items to attack, ranking them either by the confidence the deployment
reports or by a model of their own, neither of which needs the answer key
(Section~\ref{sec:e2}); and (iii) trigger this at inference
time, after the conformal certificate has been computed offline and frozen. The
goal is coverage degradation: pushing the probability that the correct answer is
\emph{not} in the prediction set from $\alpha$ to $\alpha+\Delta$. For safety-critical
uses where the certificate is the formal reliability guarantee, a $16$\,pp drop at
$\alpha=0.10$ turns a promised $90\%$ into a delivered $74\%$.

\paragraph{Clean calibration is the realistic order.}
The act-vs-escalate defeat depends on how the predictor is calibrated, and that
dependence is the point. If the threshold is calibrated on data drawn from the same
peer-pressured distribution as the test items (cal\,=\,deploy), the marginal
guarantee is honored: marginal coverage returns to $90.0\%$ and the pooled
act-on-wrong rate falls from $12.1\%$ to $2.6\%$ (Section~\ref{sec:condaware}). The
defeat appears under the realistic order instead, in which the defender calibrates offline on clean
solo data and the adversary injects conformity at test time, after the certificate is
fixed. An attacker who controls peer messages necessarily acts after calibration, so
exchangeability is broken by the attack itself, with no change to the input:
the guarantee is conditioned on an assumption the multi-agent deployment setting does
not enforce. Neither offline repair escapes this, for reasons already given:
condition-aware recalibration (Section~\ref{sec:condaware}) needs the pressure labels
the attack denies, and a tighter $\alpha$ (Section~\ref{sec:e1}) reduces autonomous
wrong actions but increases escalation while leaving half the targeted coverage gap
open.

\paragraph{Realizability and a deployment recommendation.}
The injection channel is the one recent security work has shown to be exploitable.
Multi-agent frameworks such as AutoGen~\citep{wu2023autogen} implement shared message
buses where agents read each other's intermediate outputs, and any agent with write
access can inject false peer responses. Prior work confirms this channel is
exploitable: \citet{lee2024promptinfection} show a malicious prompt can
self-replicate across such agents, and
\citet{ming2026waferqa} treat misleading peer/judge feedback as a first-class
vulnerability channel that flips correct answers. Our injected peer transcript is the
conformity-specific instance of that channel, and unlike a self-replicating payload it
need not be malicious text at all: a fluent, plausible record of agreeing peers
suffices, which is what makes the attack cheap and hard to detect. The same shape
appears in retrieval-augmented generation, debate protocols, and user-supplied context.
A full end-to-end exploit inside a deployed framework is left to future work. In the interim, a deployer choosing among candidate models should rank them by the
answer-flip stress test of Section~\ref{sec:hetero}, which transfers across task types
and so need not be run on the deployment task. Where a peer-influence channel is present, they should also suppress
autonomous action on singleton sets.

\section{Related Work}
\label{sec:related}

\paragraph{Conformal prediction and exchangeability.}
Conformal prediction returns a prediction set that covers the true label with
probability at least $1-\alpha$ under the single assumption that calibration and test
points are exchangeable~\citep{vovk2005}; we use the split variant and its standard
vocabulary~\citep{angelopoulos2021} and take its finite-sample validity theorem as
the object of attack. Two results bound what our adversary can do. \citet{tibshirani2019}
recover coverage under \emph{covariate} shift by reweighting; our shift leaves $x$
fixed, so their weight is $1$ and cannot help (Section~\ref{sec:tibshirani}).
\citet{barber2021} prove exact conditional coverage is impossible distribution-free,
so split CP never promised the subgroup guarantee our attack removes. What we report is
not that limit, however: without pressure the same subgroup sits within three points of
target, so the loss is the attack's (Section~\ref{sec:e2}).
\citet{barber2022beyond} bound the coverage gap by a distance from exchangeability, of
which our Proposition is a mechanism-specific instance. Adaptive set rules~\citep{romano2020,raps2021} and Mondrian per-group
calibration~\citep{mondrian} do not absorb the shift, which lives in the score values
rather than the set construction: Mondrian leaves three of four models below target
(Section~\ref{sec:mondrian}), and APS looks robust only because its sets degenerate to
all four options (Section~\ref{sec:e2}).

\paragraph{Conformal prediction for LLMs.}
A fast-growing line transports CP to language models and inherits the same clean
calibration we break: conformal language modeling~\citep{quach2023}, split CP for
multiple-choice QA~\citep{kumar2023}, conformal factuality~\citep{mohri2024factuality}
(whose authors note their guarantee can ``fail'' when distributions change, the
mechanism we supply), conformal abstention~\citep{abbasi2024abstention}, conformal
nucleus sampling~\citep{ravfogel2023nucleus}, and conformal risk
control~\citep{angelopoulos2024crc}. The frontier is now moving to agent pipelines
and LLM judges~\citep{gao2026agenteval,gupta2026judge}, both assuming clean
calibration, which raises the stakes of our attack.

\paragraph{Sycophancy and social conformity.}
The behavior our attack weaponizes is well documented but rarely tied to a guarantee.
It traces from \citet{asch1956}'s human conformity under a unanimous wrong majority to
LLM sycophancy at scale~\citep{perez2022discovering,sharma2023}, which worsens with
scale and resists finetuning~\citep{wei2023sycophancy}, hides behind post-hoc
rationalization~\citep{turpin2023faithfulness}, and coexists with a self-directed bias
in the same models~\citep{xu2023,sun2024}. Recent work sharpens the construct:
\citet{hao2026not} show raw flip rate overstates true conformity, which is why we
measure the coverage gap on the realized score rather than on flips; the flip rate still
orders models correctly (Section~\ref{sec:hetero}), so we use it to compare models and
the coverage gap to measure the harm; our own
\citet{qu2026easiermislead} establishes the harmful/beneficial asymmetry that predicts
the direction of our gap.

\paragraph{Multi-agent LLM systems.}
Multi-agent debate was introduced to \emph{improve} factuality on the assumption that
peer exposure is benign~\citep{du2023debate,liang2023divergent};
\citet{chen2024reconcile}'s confidence-weighted consensus is exactly what a confident
wrong majority exploits, and \citet{wu2023autogen}'s shared message bus is the
substrate that makes our injection realizable. A corrective line shows debate can
\emph{degrade} accuracy, but stops at accuracy; we add a certificate-level
consequence. At the society scale, \citet{demarzo2026conformity} show aligned agents
can settle into a misaligned consensus, the macroscopic shadow of our microscopic
attack.

\paragraph{Adversarial CP, and the closest prior work.}
Prior adversarial CP hardens against input \emph{perturbations}~\citep{adversarial_cp}.
Our adversary leaves the input untouched and acts on the communication channel, an
endogenous, calibration-clean vector those defenses cannot see (its realism is grounded
in Section~\ref{sec:threat}). The single closest paper is
\citet{wang2026debate2decision}, which uses CP to \emph{defend} multi-agent debate with
a marginal guarantee and an act-vs-escalate policy, and explicitly defers the
conditional case. We study the same defense mechanism, exhibit the conditional failure
they defer (Section~\ref{sec:e2}), and defeat their clean-calibrated gate
(Section~\ref{sec:e1}). To our knowledge this is the first adversarial, endogenous-shift
failure of the conformal guarantee under social conformity.

\section{Scope, Limitations, and Future Work}
\label{sec:limits}

We have shown that social conformity is a concrete, measurable attack surface that can
invalidate formal coverage certificates for LLMs, exploitable by anyone who controls
the peer channel, and sharpest at the conditional level, where marginal coverage can
look healthy while coverage on the targeted subgroup falls from $87\%$ to $47\%$. The effect is consistent across the models,
tasks, and three runs we tested, and its magnitude is model-dependent, governed by how far
peer pressure moves a model's correct-answer probability ($\Delta p_{\mathrm{gt}}$); we do
not claim it is universal. Our scores are token probabilities; the same collapse
appears under \citet{wang2026debate2decision}'s verbalized score, but those runs cannot
be reproduced from the released data and are not reported here.

Three boundaries sharpen the claim. Conformal prediction works exactly as advertised
under exchangeability, the assumption a multi-agent deployment does not
enforce. Vulnerability is uneven: Gemma-9B is much more robust than Qwen-7B, at least
marginally. And condition-aware recalibration does fix it, but needs pressure-labeled
calibration data a defender usually cannot obtain in advance, and buys the fix by
widening the sets until the layer escalates nearly everything
(Section~\ref{sec:condaware}).

Several directions remain open. The most important is a practical \emph{defense}:
detecting pressured items, selective abstention, or running the conformity stress test
online as a detector. Candidate repairs worth testing include mixture or augmented calibration
over synthetic peer contexts (approximating the pressured distribution without per-item
labels), worst-case conditioning, and online or drift-detection CP. Cross-conformal,
jackknife+, and e-value variants could tighten finite-sample slack, but none restore
exchangeability under an adaptive adversary. Sharpening the threat itself, a systematic
\emph{attacker dial} over the number of wrong peers, \emph{safety-domain data}, and an
\emph{end-to-end exploit} inside a deployed framework are natural next steps.

The broader lesson is simple: clean calibration certifies how a model behaves alone,
not under pressure, so in a multi-agent LLM system a certificate must certify the social
context, not just the input.

\subsubsection*{Acknowledgments}

This work used Jetstream2 at Indiana University through ACCESS allocation CIS260254
from the Advanced Cyberinfrastructure Coordination Ecosystem: Services \& Support
(ACCESS) program, which is supported by U.S. National Science Foundation grants
\#2138259, \#2138286, \#2138307, \#2137603, and \#2138296. Results were also obtained
using the Chameleon testbed, supported by the National Science Foundation. We thank the
Jetstream2, ACCESS, and Chameleon support teams for the computational infrastructure
used in this work.

\bibliographystyle{iclr2026_conference}
\bibliography{refs}

\begin{thebibliography}{34}
\providecommand{\natexlab}[1]{#1}
\providecommand{\url}[1]{\texttt{#1}}
\expandafter\ifx\csname urlstyle\endcsname\relax
  \providecommand{\doi}[1]{doi: #1}\else
  \providecommand{\doi}{doi: \begingroup \urlstyle{rm}\Url}\fi

\bibitem[Abbasi~Yadkori et~al.(2024)Abbasi~Yadkori, Kuzborskij, Stutz,
  Gy{\"o}rgy, Fisch, Doucet, Beloshapka, Weng, Yang, Szepesv{\'a}ri, Cemgil,
  and Tomasev]{abbasi2024abstention}
Yasin Abbasi~Yadkori, Ilja Kuzborskij, David Stutz, Andr{\'a}s Gy{\"o}rgy, Adam
  Fisch, Arnaud Doucet, Iuliya Beloshapka, Wei-Hung Weng, Yao~Yuan Yang, Csaba
  Szepesv{\'a}ri, Ali~Taylan Cemgil, and Nenad Tomasev.
\newblock Mitigating {LLM} hallucinations via conformal abstention, 2024.
\newblock URL \url{https://arxiv.org/abs/2405.01563}.

\bibitem[Angelopoulos \& Bates(2021)Angelopoulos and Bates]{angelopoulos2021}
Anastasios~N. Angelopoulos and Stephen Bates.
\newblock A gentle introduction to conformal prediction and distribution-free
  uncertainty quantification.
\newblock \emph{arXiv preprint arXiv:2107.07511}, 2021.

\bibitem[Angelopoulos et~al.(2021)Angelopoulos, Bates, Jordan, and
  Malik]{raps2021}
Anastasios~N. Angelopoulos, Stephen Bates, Michael~I. Jordan, and Jitendra
  Malik.
\newblock Uncertainty sets for image classifiers using conformal prediction.
\newblock In \emph{ICLR}, 2021.

\bibitem[Angelopoulos et~al.(2024)Angelopoulos, Bates, Fisch, Lei, and
  Schuster]{angelopoulos2024crc}
Anastasios~N. Angelopoulos, Stephen Bates, Adam Fisch, Lihua Lei, and Tal
  Schuster.
\newblock Conformal risk control.
\newblock In \emph{ICLR}, 2024.

\bibitem[Asch(1956)]{asch1956}
Solomon~E. Asch.
\newblock Studies of independence and conformity: {I}. {A} minority of one
  against a unanimous majority.
\newblock \emph{Psychological Monographs: General and Applied}, 70\penalty0
  (9):\penalty0 1--70, 1956.
\newblock \doi{10.1037/h0093718}.

\bibitem[Barber et~al.(2021)Barber, Candes, Ramdas, and Tibshirani]{barber2021}
Rina~Foygel Barber, Emmanuel~J. Candes, Aaditya Ramdas, and Ryan~J. Tibshirani.
\newblock Limits of distribution-free conditional predictive inference.
\newblock \emph{Information and Inference}, 10\penalty0 (2):\penalty0 455--482,
  2021.
\newblock \doi{10.1093/imaiai/iaaa017}.

\bibitem[Barber et~al.(2023)Barber, Cand{\`e}s, Ramdas, and
  Tibshirani]{barber2022beyond}
Rina~Foygel Barber, Emmanuel~J. Cand{\`e}s, Aaditya Ramdas, and Ryan~J.
  Tibshirani.
\newblock Conformal prediction beyond exchangeability.
\newblock \emph{The Annals of Statistics}, 51\penalty0 (2):\penalty0 816--845,
  2023.
\newblock \doi{10.1214/23-AOS2276}.
\newblock arXiv:2202.13415.

\bibitem[Chen et~al.(2024)Chen, Saha, and Bansal]{chen2024reconcile}
Justin Chih-Yao Chen, Swarnadeep Saha, and Mohit Bansal.
\newblock {ReConcile}: Round-table conference improves reasoning via consensus
  among diverse {LLM}s.
\newblock In \emph{ACL}, 2024.
\newblock \doi{10.18653/v1/2024.acl-long.381}.
\newblock URL \url{https://arxiv.org/abs/2309.13007}.

\bibitem[De~Marzo et~al.(2026)De~Marzo, Bellina, Castellano, Priesemann, and
  Garcia]{demarzo2026conformity}
Giordano De~Marzo, Alessandro Bellina, Claudio Castellano, Viola Priesemann,
  and David Garcia.
\newblock Conformity generates collective misalignment in {AI} agents
  societies, 2026.
\newblock URL \url{https://arxiv.org/abs/2605.10721}.

\bibitem[Du et~al.(2024)Du, Li, Torralba, Tenenbaum, and
  Mordatch]{du2023debate}
Yilun Du, Shuang Li, Antonio Torralba, Joshua~B. Tenenbaum, and Igor Mordatch.
\newblock Improving factuality and reasoning in language models through
  multiagent debate.
\newblock In \emph{ICML}, 2024.

\bibitem[Gao et~al.(2026)Gao, Wang, and Yu]{gao2026agenteval}
Yuxuan Gao, Megan Wang, and Yi~Ling Yu.
\newblock Distribution-free uncertainty quantification for continuous {AI}
  agent evaluation, 2026.
\newblock URL \url{https://arxiv.org/abs/2605.19779}.

\bibitem[Ghosh et~al.(2023)Ghosh, Shi, Belkhouja, Yan, Doppa, and
  Jones]{adversarial_cp}
Subhankar Ghosh, Yuanjie Shi, Taha Belkhouja, Yan Yan, Jana Doppa, and Brian
  Jones.
\newblock Probabilistically robust conformal prediction.
\newblock In \emph{Uncertainty in Artificial Intelligence (UAI)}, 2023.
\newblock arXiv:2307.16360.

\bibitem[Gupta \& Kumar(2026)Gupta and Kumar]{gupta2026judge}
Manan Gupta and Dhruv Kumar.
\newblock Diagnosing {LLM} judge reliability: Conformal prediction sets and
  transitivity violations, 2026.
\newblock URL \url{https://arxiv.org/abs/2604.15302}.

\bibitem[Hao et~al.(2026)Hao, Wu, Qiu, Xiao, Xu, Zheng, and Qin]{hao2026not}
Xiqi Hao, Zengqing Wu, Yu-Xuan Qiu, Chuan Xiao, Ruiqi Xu, Shuyuan Zheng, and
  Jianbin Qin.
\newblock Not all flips are conformity: Decomposing stance convergence in
  multi-agent {LLM} debate, 2026.
\newblock URL \url{https://arxiv.org/abs/2606.00820}.

\bibitem[Kumar et~al.(2023)Kumar, Lu, Gupta, Palepu, Bellamy, Raskar, and
  Beam]{kumar2023}
Bhawesh Kumar, Charlie Lu, Gauri Gupta, Anil Palepu, David Bellamy, Ramesh
  Raskar, and Andrew Beam.
\newblock Conformal prediction with large language models for multi-choice
  question answering.
\newblock In \emph{ICML 2023 Workshop on Neural Conversational AI (TEACH)},
  2023.
\newblock URL \url{https://arxiv.org/abs/2305.18404}.

\bibitem[Lee \& Tiwari(2024)Lee and Tiwari]{lee2024promptinfection}
Donghyun Lee and Mo~Tiwari.
\newblock Prompt infection: {LLM}-to-{LLM} prompt injection within multi-agent
  systems, 2024.
\newblock URL \url{https://arxiv.org/abs/2410.07283}.

\bibitem[Liang et~al.(2024)Liang, He, Jiao, Wang, Wang, Wang, Yang, Shi, and
  Tu]{liang2023divergent}
Tian Liang, Zhiwei He, Wenxiang Jiao, Xing Wang, Yan Wang, Rui Wang, Yujiu
  Yang, Shuming Shi, and Zhaopeng Tu.
\newblock Encouraging divergent thinking in large language models through
  multi-agent debate.
\newblock In \emph{EMNLP}, 2024.
\newblock \doi{10.18653/v1/2024.emnlp-main.992}.
\newblock URL \url{https://arxiv.org/abs/2305.19118}.

\bibitem[Ming et~al.(2025)Ming, Ke, Nguyen, Wang, and Joty]{ming2026waferqa}
Yifei Ming, Zixuan Ke, Xuan-Phi Nguyen, Jiayu Wang, and Shafiq Joty.
\newblock Helpful agent meets deceptive judge: Understanding vulnerabilities in
  agentic workflows, 2025.
\newblock URL \url{https://arxiv.org/abs/2506.03332}.

\bibitem[Mohri \& Hashimoto(2024)Mohri and Hashimoto]{mohri2024factuality}
Christopher Mohri and Tatsunori Hashimoto.
\newblock Language models with conformal factuality guarantees, 2024.
\newblock URL \url{https://arxiv.org/abs/2402.10978}.
\newblock ICML 2024.

\bibitem[Perez et~al.(2023)Perez, Ringer, Lukosiute, Nguyen, Chen,
  et~al.]{perez2022discovering}
Ethan Perez, Sam Ringer, Kamile Lukosiute, Karina Nguyen, Edwin Chen, et~al.
\newblock Discovering language model behaviors with model-written evaluations.
\newblock In \emph{Findings of ACL}, 2023.
\newblock \doi{10.18653/v1/2023.findings-acl.847}.
\newblock arXiv:2212.09251.

\bibitem[Qu et~al.(2026)Qu, Fu, and Hu]{qu2026easiermislead}
Jiaming Qu, Lucheng Fu, and Yibo Hu.
\newblock Easier to mislead than to correct: Harmful and beneficial revision in
  {LLM} conformity, 2026.
\newblock URL \url{https://arxiv.org/abs/2606.01637}.

\bibitem[Quach et~al.(2024)Quach, Fisch, Schuster, Yala, Sohn, Jaakkola, and
  Barzilay]{quach2023}
Victor Quach, Adam Fisch, Tal Schuster, Adam Yala, Jae~Ho Sohn, Tommi~S.
  Jaakkola, and Regina Barzilay.
\newblock Conformal language modeling.
\newblock In \emph{ICLR}, 2024.
\newblock URL \url{https://arxiv.org/abs/2306.10193}.

\bibitem[Ravfogel et~al.(2023)Ravfogel, Goldberg, and
  Goldberger]{ravfogel2023nucleus}
Shauli Ravfogel, Yoav Goldberg, and Jacob Goldberger.
\newblock Conformal nucleus sampling.
\newblock In \emph{Findings of ACL}, 2023.
\newblock \doi{10.18653/v1/2023.findings-acl.3}.

\bibitem[Romano et~al.(2020)Romano, Sesia, and Cand{\`e}s]{romano2020}
Yaniv Romano, Matteo Sesia, and Emmanuel Cand{\`e}s.
\newblock Classification with valid and adaptive coverage.
\newblock In \emph{NeurIPS}, 2020.

\bibitem[Sharma et~al.(2024)Sharma, Tong, Korbak, Duvenaud, Askell, Bowman,
  Cheng, Durmus, Hatfield-Dodds, Johnston, Kravec, Maxwell, McCandlish,
  Ndousse, Rausch, Schiefer, Yan, Zhang, and Perez]{sharma2023}
Mrinank Sharma, Meg Tong, Tomasz Korbak, David Duvenaud, Amanda Askell,
  Samuel~R. Bowman, Newton Cheng, Esin Durmus, Zac Hatfield-Dodds, Scott~R.
  Johnston, Shauna Kravec, Timothy Maxwell, Sam McCandlish, Kamal Ndousse,
  Oliver Rausch, Nicholas Schiefer, Da~Yan, Miranda Zhang, and Ethan Perez.
\newblock Towards understanding sycophancy in language models.
\newblock In \emph{ICLR}, 2024.
\newblock URL \url{https://arxiv.org/abs/2310.13548}.

\bibitem[Sun et~al.(2023)Sun, Shen, Zhou, Zhang, Chen, Cox, Yang, and
  Gan]{sun2024}
Zhiqing Sun, Yikang Shen, Qinhong Zhou, Hongxin Zhang, Zhenfang Chen, David
  Cox, Yiming Yang, and Chuang Gan.
\newblock Principle-driven self-alignment of language models from scratch with
  minimal human supervision.
\newblock \emph{NeurIPS}, 2023.
\newblock \doi{10.52202/075280-0115}.

\bibitem[Tibshirani et~al.(2019)Tibshirani, Barber, Candes, and
  Ramdas]{tibshirani2019}
Ryan~J. Tibshirani, Rina~Foygel Barber, Emmanuel~J. Candes, and Aaditya Ramdas.
\newblock Conformal prediction under covariate shift.
\newblock In \emph{NeurIPS}, 2019.

\bibitem[Turpin et~al.(2023)Turpin, Michael, Perez, and
  Bowman]{turpin2023faithfulness}
Miles Turpin, Julian Michael, Ethan Perez, and Samuel~R. Bowman.
\newblock Language models don't always say what they think: Unfaithful
  explanations in chain-of-thought prompting.
\newblock In \emph{NeurIPS}, 2023.

\bibitem[Vovk(2012)]{mondrian}
Vladimir Vovk.
\newblock Conditional validity of inductive conformal predictors.
\newblock In \emph{ACML}, 2012.
\newblock \doi{10.1007/s10994-013-5355-6}.

\bibitem[Vovk et~al.(2005)Vovk, Gammerman, and Shafer]{vovk2005}
Vladimir Vovk, Alex Gammerman, and Glenn Shafer.
\newblock \emph{Algorithmic Learning in a Random World}.
\newblock Springer, 2005.
\newblock \doi{10.1007/b106715}.

\bibitem[Wang et~al.(2026)Wang, Xie, Wang, Gao, Yang, Li, Qiu, Han, Qiu, Huang,
  Zhu, and Woo]{wang2026debate2decision}
Mengdie~Flora Wang, Haochen Xie, Guanghui Wang, Aijing Gao, Guang Yang, Ziyuan
  Li, Qucy~Wei Qiu, Fangwei Han, Hengzhi Qiu, Yajing Huang, Bing Zhu, and
  Jae~Oh Woo.
\newblock From debate to decision: Conformal social choice for safe multi-agent
  deliberation, 2026.
\newblock URL \url{https://arxiv.org/abs/2604.07667}.

\bibitem[Wei et~al.(2023)Wei, Huang, Lu, Zhou, and Le]{wei2023sycophancy}
Jerry Wei, Da~Huang, Yifeng Lu, Denny Zhou, and Quoc~V. Le.
\newblock Simple synthetic data reduces sycophancy in large language models.
\newblock \emph{arXiv preprint arXiv:2308.03958}, 2023.

\bibitem[Wu et~al.(2023)Wu, Bansal, Zhang, Wu, Li, Zhu, Jiang, Zhang, Zhang,
  Liu, Awadallah, White, Burger, and Wang]{wu2023autogen}
Qingyun Wu, Gagan Bansal, Jieyu Zhang, Yiran Wu, Beibin Li, Erkang Zhu,
  Li~Jiang, Xiaoyun Zhang, Shaokun Zhang, Jiale Liu, Ahmed~Hassan Awadallah,
  Ryen~W. White, Doug Burger, and Chi Wang.
\newblock {AutoGen}: Enabling next-gen {LLM} applications via multi-agent
  conversation.
\newblock \emph{arXiv preprint arXiv:2308.08155}, 2023.

\bibitem[Xu et~al.(2024)Xu, Zhu, Zhao, Pan, Li, and Wang]{xu2023}
Wenda Xu, Guanglei Zhu, Xuandong Zhao, Liangming Pan, Lei Li, and William~Yang
  Wang.
\newblock Pride and prejudice: {LLM} amplifies self-bias in self-refinement.
\newblock In \emph{Proceedings of the 62nd Annual Meeting of the Association
  for Computational Linguistics (Volume 1: Long Papers)}, 2024.
\newblock \doi{10.18653/v1/2024.acl-long.826}.

\end{thebibliography}

\clearpage
\appendix

\noindent\textbf{Appendix roadmap.}
Appendix~\ref{app:robgen} (\textbf{robustness \& generalization}) collects the
per-condition, Mondrian, decomposition, adaptive-adversary, and per-task evidence;
Appendix~\ref{app:condfull} gives the \textbf{full conditional-coverage table};
Appendix~\ref{app:probes} holds an \textbf{extra probe} (the pressure
dose-response); Appendix~\ref{app:screen} \textbf{validates the pre-deployment stress
test}; Appendix~\ref{app:qwensize} reports the \textbf{Qwen size-scaling} probe;
Appendix~\ref{app:numbers} collects \textbf{canonical numbers and reproducibility};
and Appendix~\ref{app:proof} gives the \textbf{proof of Proposition~\ref{prop:coverage}}.

\section{Robustness and generalization}
\label{app:robgen}

These figures and tables support the main-text claims of
Sections~\ref{sec:marginal}, \ref{sec:e2}, \ref{sec:mondrian}, \ref{sec:hetero},
and \ref{sec:perdataset}.

\paragraph{Experimental protocol (full detail).}
All distributions were collected on Jetstream2 A100 GPUs. We use three seeds: the
questions and the models' solo (round-1) answers are held fixed, and only the
fabricated peer transcripts are resampled, so reported spreads over seeds measure
sensitivity to the injected transcript. Seed~0 covers 4 models $\times$ 2 datasets
(ARC-Challenge and TruthfulQA), $1{,}200$ question--model pairs, and additionally holds
Qwen2.5-14B and Qwen2.5-32B on the same two datasets, from which the frontier row of
Section~\ref{sec:e1} is taken; seeds~1 and~2 each cover 6 models $\times$ 7 task types,
$\approx 4{,}000$--$10{,}500$ pairs. The models
are Qwen2.5-7B-Instruct, Llama-3.1-8B-Instruct, Mistral-7B-Instruct-v0.3, and
Gemma-2-9B-IT (all seeds), plus Qwen2.5-1.5B and Qwen2.5-3B (seeds~1--2).

\paragraph{Peer transcripts.}
Each condition inserts six one-line peer turns, joined by blank lines, after the
question. Speaker names are drawn without replacement from a fixed pool. Wording is
sampled from ten interchangeable templates: ``\{name\}: I think this answer is
\{label\}.'', ``\ldots{} My guess is \ldots'', ``\ldots{} I am leaning toward
\ldots'', ``\ldots{} I would go with \ldots'', and six more of the same form. In the
$+$authority conditions exactly one of the six speakers carries an institutional role
drawn from \{lab supervisor, committee chair, program director, operations lead\}; the
other five are unchanged. The mixed and unanimous-correct conditions use the same
templates with different label assignments. The transcript for a given (item,
condition) is resampled per experiment seed while the question and the model's solo
answer stay fixed, which is what the seed spread measures.

\begin{figure}[h]
  \centering
  \includegraphics[width=0.78\textwidth]{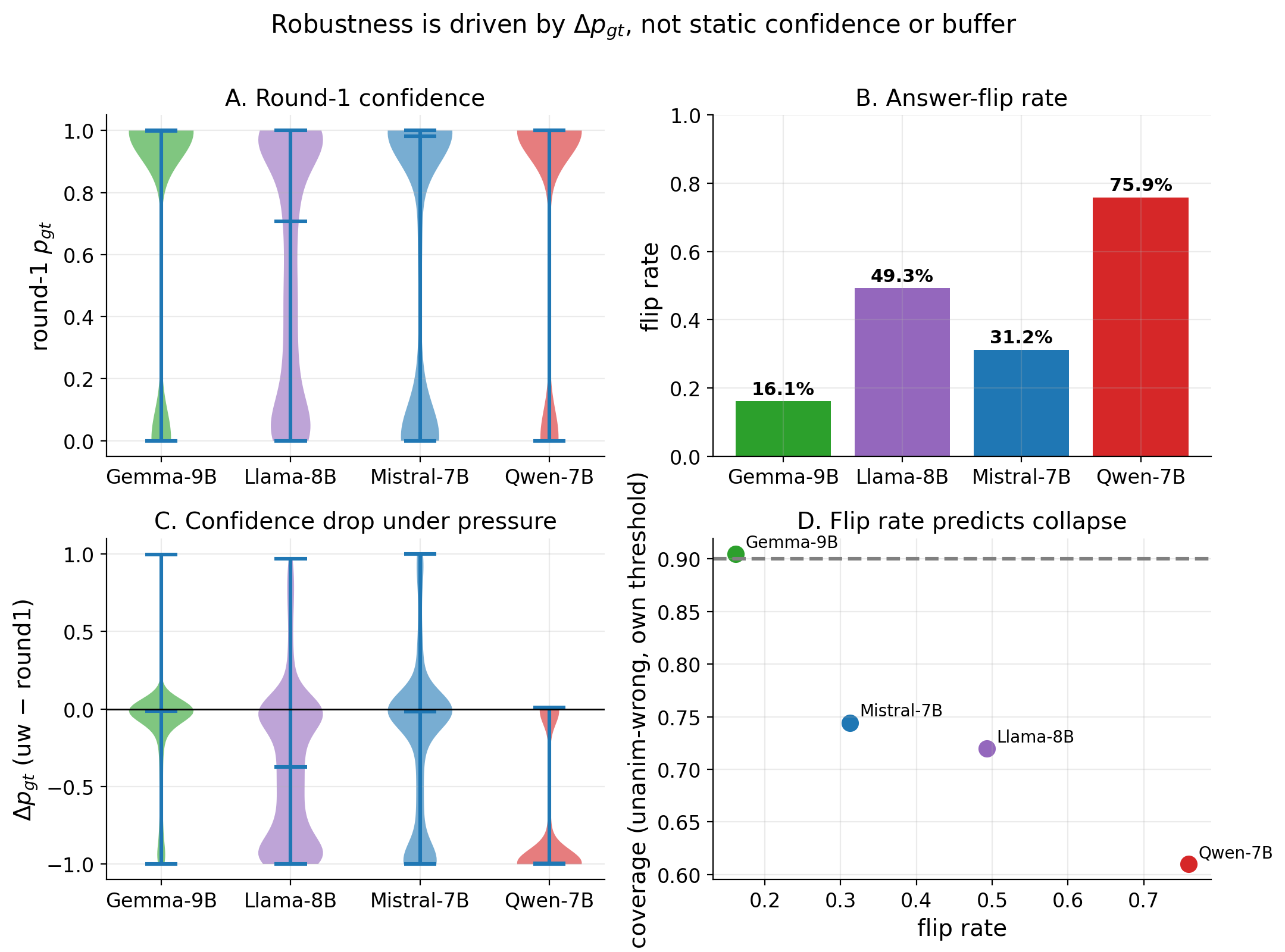}
  \caption{Robustness decomposition across four models. Panel A: round-1 confidence
    (Gemma and Qwen are similar). Panel B: answer-flip rate (Gemma $16\%$, Qwen
    $76\%$). Panel C: $p_{\mathrm{gt}}$ drop under unanimous-wrong pressure (Gemma
    $-0.15$, Qwen $-0.75$). Panel D: answer-flip rate vs.\ coverage under
    pressure, each model on its own threshold so that the shared-calibration artifact
    of Section~\ref{sec:mondrian} does not enter. The robustness gap is driven by
    $\Delta p_{\mathrm{gt}}$, not static confidence or buffer.}
  \label{fig:robustness}
\end{figure}

\begin{figure}[h]
  \centering
  \includegraphics[width=0.74\textwidth]{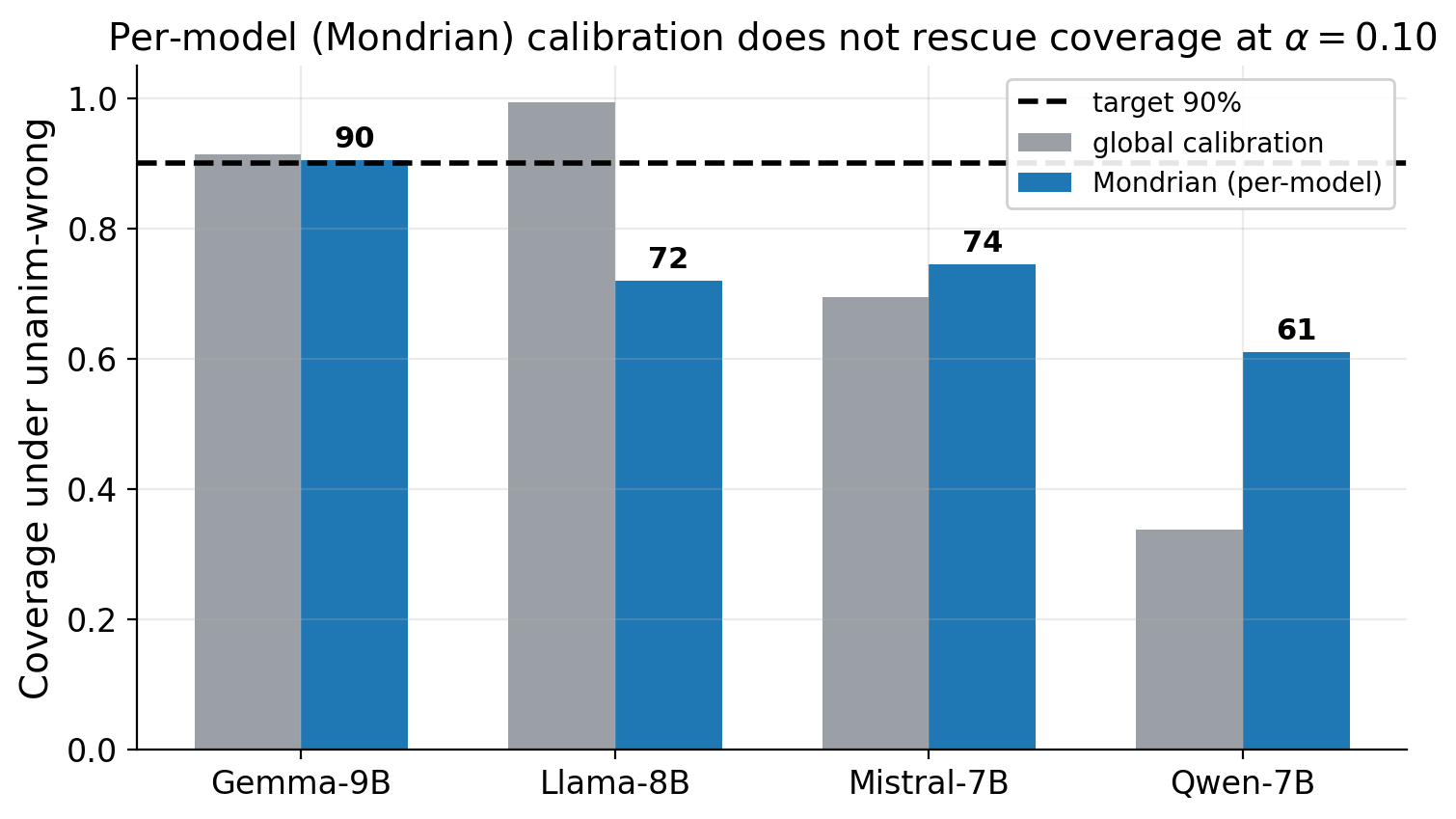}
  \caption{Per-model (Mondrian) calibration does not rescue coverage at
    $\alpha=0.10$. It removes the global-pooling artifact (Llama's apparent immunity)
    but three of four models still fail under unanimous-wrong pressure (Qwen $61\%$,
    Llama $\approx72\%$, Mistral $\approx74\%$; only Gemma stays near $90\%$). At
    $\alpha=0.05$ Mondrian largely rescues the vulnerable models ($88$--$92\%$),
    so the persistent failure is the $\alpha=0.10$ finding shown here.}
  \label{fig:mondrian}
\end{figure}

\begin{figure}[h]
  \centering
  \includegraphics[width=0.95\textwidth]{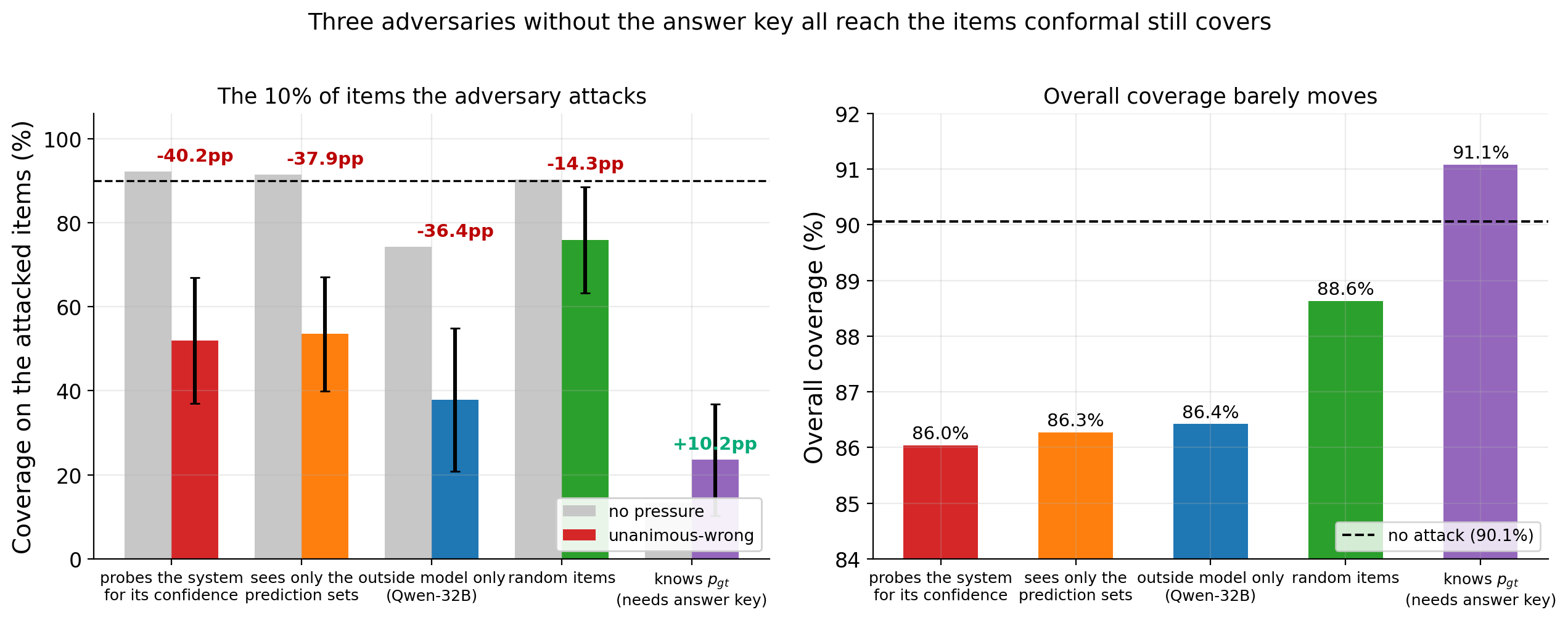}
  \caption{Targeting without the answer key. Attacking $10\%$ of items, three
    adversaries with no access to the correct answers cost $-36$ to $-40$\,pp on the
    items they choose, against $-14.3$\,pp for random targeting. Ranking by
    $p_{\mathrm{gt}}$ instead selects the decile the threshold already excludes, where
    coverage is $13.4\%$ to begin with and the attack raises it. Overall coverage moves
    by at most $4$\,pp throughout (right).}
  \label{fig:adversary}
\end{figure}

\begin{table}[h]
\centering
\caption{Robustness decomposition (three-seed pooled). ``Buffer'' $=
\mathrm{mean}(p_{\mathrm{gt}}^{(r1)}-(1-\hat\tau))$ for covered calibration items;
$\Delta p_{\mathrm{gt}}$ is the mean change in ground-truth probability under
unanimous-wrong pressure. The driver of collapse is $\Delta p_{\mathrm{gt}}$, not
static confidence or buffer. The coverage column uses the shared threshold, which is
why Llama reads as immune; on per-model thresholds the same four models read $61$, $72$,
$74$ and $90\%$ (Section~\ref{sec:hetero}).}
\label{tab:robustness}
\small
\begin{tabular}{lcccccc}
\toprule
Model & Flip rate & $\bar{p}_{\mathrm{gt}}^{(r1)}$ & Median $p_{\mathrm{gt}}^{(r1)}$
      & $\Delta p_{\mathrm{gt}}$ & Buffer & Coverage (UW) \\
\midrule
Gemma-9B  & \textcolor{ok}{16.2\%}  & 0.753 & 0.999 & \textcolor{ok}{$-$0.151} & 0.818 & \textcolor{ok}{90.6\%} \\
Mistral-7B & 31.2\% & 0.612 & 0.981 & $-$0.226 & 0.779 & \textcolor{collapse}{68.4\%} \\
Llama-8B  & 49.3\% & 0.583 & 0.707 & $-$0.384 & 0.555 & 99.3\% (artifact) \\
\rowcolor{red!8}
Qwen-7B   & \textcolor{collapse}{75.8\%}  & 0.765 & 1.000 & \textcolor{collapse}{$-$0.747} & 0.862 & \textcolor{collapse}{37.5\%} \\
\bottomrule
\end{tabular}
\end{table}

\begin{table}[h]
\centering
\caption{Coverage under unanimous-wrong pressure per task type (seeds~1--2, 6 models
pooled, global calibration, $\alpha=0.10$). The four BIG-Bench Hard subtasks
(geometric shapes, logical deduction, temporal sequences, tracking) are bundled into
one ``BBH'' row. Every task type shows substantial collapse. The gap column is measured against
the $90\%$ target. TruthfulQA's round-1 coverage ($85.5\%$) is already below nominal
with no pressure (a known truth-direction confound), so part of its $-19.3$\,pp is not
the conformity effect; against its own clean baseline the pressure costs $-14.8$\,pp.
The headline numbers elsewhere, computed against the calibrated threshold, are
unaffected.}
\label{tab:perdataset}
\small
\begin{tabular}{lccc}
\toprule
Dataset & Round-1 coverage & Unanimous-wrong & Gap \\
\midrule
ARC-Challenge & 96.6\% & \textcolor{collapse}{84.7\%} & $-$5.3pp \\
BBH (multi-task) & 90.3\% & \textcolor{collapse}{65.7\%} & $-$24.3pp \\
MMLU-Pro & 87.7\% & \textcolor{collapse}{62.3\%} & $-$27.7pp \\
TruthfulQA & 85.5\% & \textcolor{collapse}{70.7\%} & $-$19.3pp \\
\bottomrule
\end{tabular}
\end{table}

\section{Full conditional-coverage results}
\label{app:condfull}

Table~\ref{tab:conditional} gives the full per-model conditional-coverage numbers
summarized by Figure~\ref{fig:e2}.

\paragraph{Why the targeted band, and not the bottom $k\%$.}
Table~\ref{tab:strata} splits the pooled test set into strata of round-1 confidence.
The attack removes coverage everywhere above the calibrated threshold, most sharply in
the decile just above it, and the effect decays monotonically with confidence. Below the
threshold the pattern reverses: coverage \emph{rises} from $13.4\%$ to $23.6\%$. This is
a floor effect, not protection. The median clean $p_{\mathrm{gt}}$ there is $0.006$, so
the predictor is already confidently wrong and almost nothing remains to be lost; peer
pressure moves mass off the predictor's own wrong answer toward the peers', and for
$63\%$ of these items some mass returns to the correct option. The bottom-$20\%$ band
averages the two regimes and so understates the attack, and the bottom-$10\%$ column of
Table~\ref{tab:conditional} inherits the reversal, which is why we report the band above
the threshold instead.

\begin{table}[h]
\centering
\caption{Coverage by stratum of round-1 confidence (pooled predictor, seed-0,
$\alpha=0.10$, mean over $2{,}000$ splits). The calibrated threshold falls at the tenth
percentile.}
\label{tab:strata}
\small
\begin{tabular}{lccc}
\toprule
Stratum (round-1 $p_{\mathrm{gt}}$ percentile) & No pressure & Unanimous wrong & Change \\
\midrule
$0$--$10$ (below threshold) & 13.4\% & 23.6\% & \textcolor{ok}{$+$10.2\,pp} \\
\rowcolor{red!8}
$10$--$20$ (the targeted band) & 87.3\% & \textcolor{collapse}{47.4\%} & \textcolor{collapse}{$-$39.9\,pp} \\
$20$--$40$ & 100.0\% & 70.7\% & \textcolor{collapse}{$-$29.3\,pp} \\
$40$--$70$ & 100.0\% & 84.3\% & $-$15.7\,pp \\
$70$--$100$ & 100.0\% & 97.3\% & $-$2.7\,pp \\
\bottomrule
\end{tabular}
\end{table}

\begin{table}[h]
\centering
\caption{Marginal vs.\ conditional coverage under unanimous-wrong pressure, LAC
split-conformal calibrated on clean round-1 data ($\alpha=0.10$, target $90\%$; mean
$\pm$ std over $2{,}000$ splits). \emph{Band} is the headline subgroup: the decile of
round-1 $p_{\mathrm{gt}}$ immediately above the calibrated threshold. The bottom-$k\%$
columns are given for reference; they straddle the threshold and are discussed below.}
\label{tab:conditional}
\small
\setlength{\tabcolsep}{4pt}
\begin{tabular}{lcccccc}
\toprule
 & \multicolumn{2}{c}{Marginal} & \multicolumn{2}{c}{Band} & \multicolumn{2}{c}{Conditional} \\
\cmidrule(lr){2-3}\cmidrule(lr){4-5}\cmidrule(lr){6-7}
Unit & Clean & Unan.-wrong & Clean & Unan.-wrong & Bottom-20\% & Bottom-10\% \\
\midrule
\rowcolor{red!8}
\textbf{POOLED} & 90.1\%$\pm$3.3 & 75.7\%$\pm$7.0  & 87.3\%$\pm$20.2 & \textcolor{collapse}{47.4\%$\pm$21.0} & \textcolor{collapse}{\textbf{35.5\%}$\pm$15.3} & \textcolor{collapse}{\textbf{23.6\%}$\pm$13.2} \\
Qwen-7B   & 90.0\%$\pm$3.2 & 60.1\%$\pm$16.5  & 87.0\%$\pm$20.6 & \textcolor{collapse}{36.3\%$\pm$14.3} & \textcolor{collapse}{26.2\%$\pm$14.0} & \textcolor{collapse}{16.1\%$\pm$16.5} \\
Llama-8B  & 90.1\%$\pm$3.3 & 73.0\%$\pm$11.9  & 87.7\%$\pm$20.1 & \textcolor{collapse}{69.0\%$\pm$15.6} & 58.2\%$\pm$15.2 & 47.3\%$\pm$17.8 \\
Mistral-7B & 90.1\%$\pm$3.5 & 81.2\%$\pm$2.8  & 86.7\%$\pm$21.0 & \textcolor{collapse}{76.3\%$\pm$8.8} & 78.9\%$\pm$5.4 & 81.4\%$\pm$8.5 \\
Gemma-9B  & 90.0\%$\pm$3.4 & 90.1\%$\pm$3.6  & 86.6\%$\pm$21.0 & \textcolor{collapse}{70.6\%$\pm$16.3} & 53.8\%$\pm$14.2 & 37.0\%$\pm$15.1 \\
\bottomrule
\end{tabular}
\end{table}

\section{Extra probe: pressure dose-response}
\label{app:probes}

The check below corroborates the main result but is supporting rather than central: the
dose-response curve is an illustration on a synthetic pressure axis rather than an
attacker-controlled dial. The mechanism itself rests on the Proposition of
Section~\ref{sec:prop}, which operates at the level of individual scores.

\subsection{How much pressure is needed to break coverage?}
\label{sec:dose}

\begin{figure}[h]
  \centering
  \includegraphics[width=0.72\textwidth]{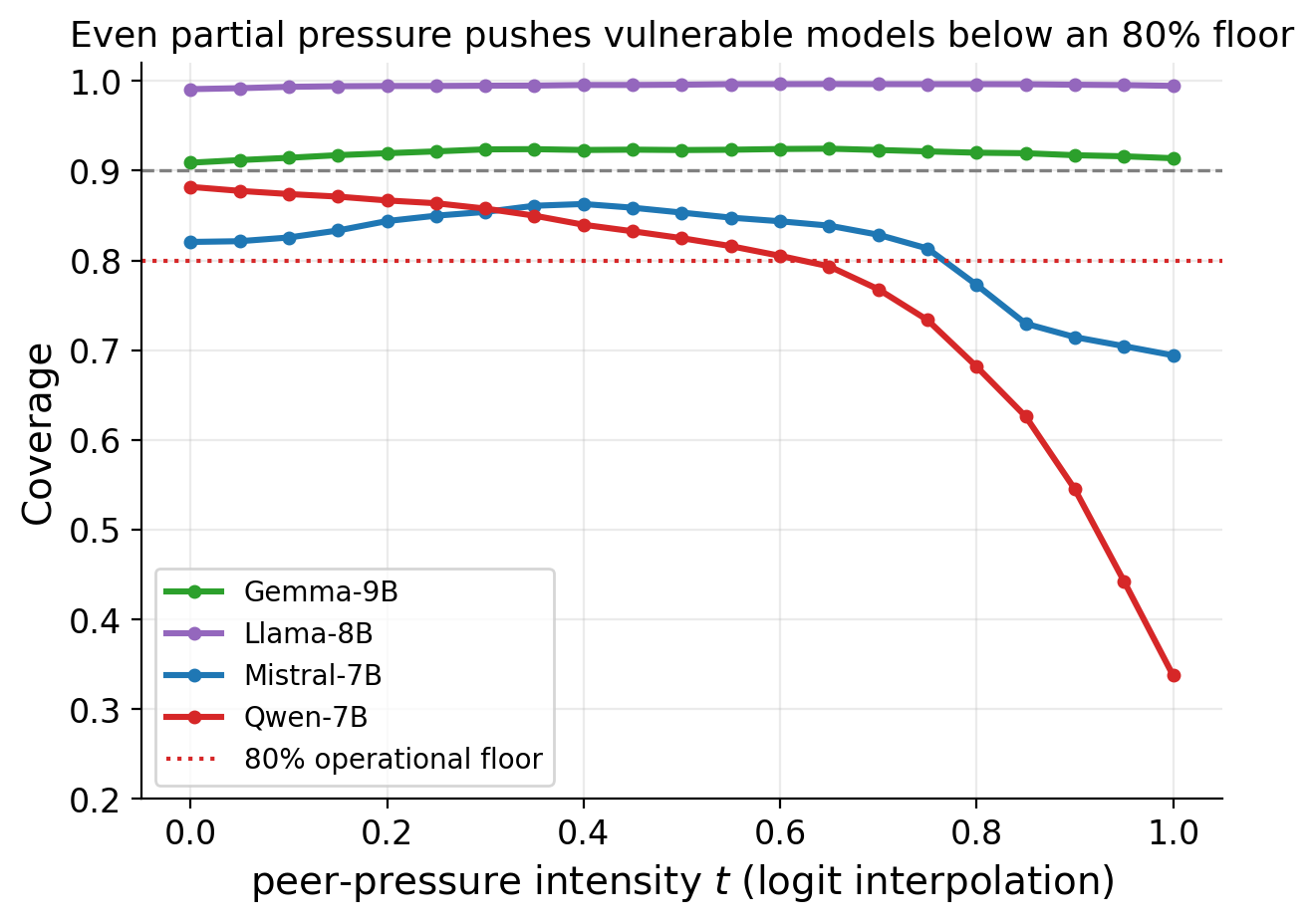}
  \caption{Coverage vs.\ peer-pressure intensity (logit interpolation). Qwen-7B falls
    below $80\%$ at $t\approx0.65$, Mistral-7B at $t\approx0.80$; Gemma-9B and
    Llama-8B (global-calibration artifact) never break $80\%$.}
  \label{fig:dose}
\end{figure}

We model a dose-response curve by interpolating on the logit scale between round-1
($t=0$, no peers) and unanimous-wrong ($t=1$, full pressure),
\[
  p_{\mathrm{gt}}(t) = \sigma\!\left(
    (1-t)\,\mathrm{logit}(p_{\mathrm{gt}}^{(r1)})
    + t\,\mathrm{logit}(p_{\mathrm{gt}}^{(\mathrm{uw})})\right),
\]
and measure coverage at each $t$ (Figure~\ref{fig:dose}). Qwen's certificate breaks
($<80\%$ coverage) at $t\approx0.65$ and Mistral's at $t\approx0.80$,
while Gemma and Llama (global-calibration artifact) never break $80\%$ on this
synthetic axis. We present this as an illustration of how the coverage gap grows
smoothly with the magnitude of the score shift, not as an attacker dial: $t$ is a
logit-space interpolation, not a quantity an attacker directly controls (the real
dial, the number of wrong peers, is left to future work). The curve confirms that
even partial pressure suffices to push the vulnerable models below an operational
coverage floor.

\section{Validating the pre-deployment stress test}
\label{app:screen}

Section~\ref{sec:hetero} recommends ranking candidate models by an answer-flip stress
test before deployment. Two checks support it. Coverage is always measured on per-model thresholds, so
none of these numbers carry the shared-calibration artifact of
Section~\ref{sec:mondrian}; every model is calibrated to $89.9$--$90.2\%$ clean
coverage, and the Monte-Carlo error over $2{,}000$ splits is at most $0.4$\,pp, well
below the spreads being ranked.

First, the stress test is not replaceable by anything static. Table~\ref{tab:screenpred}
ranks four candidate predictors against pressured coverage. The two dynamic ones, which
require actually applying peer pressure, rank it; the two static ones do not, and their
sign flips between regimes.

\begin{table}[h]
\centering
\caption{Spearman rank correlation between each predictor and pressured coverage, over
the six models of each regime. Negative is the correct direction for the dynamic
quantities, positive for the static ones.}
\label{tab:screenpred}
\small
\begin{tabular}{lcc}
\toprule
Predictor & ARC $+$ TruthfulQA & Seven task types \\
\midrule
\rowcolor{green!7}
Answer-flip rate & \textcolor{ok}{$-0.89$} & \textcolor{ok}{$-0.71$} \\
$|\Delta p_{\mathrm{gt}}|$ & \textcolor{ok}{$-0.89$} & $-0.49$ \\
\midrule
Mean solo $p_{\mathrm{gt}}$ & $-0.20$ & $+0.37$ \\
Clean accuracy & $-0.20$ & $+0.37$ \\
\bottomrule
\end{tabular}
\end{table}

Second, the ranking holds out of sample. Because the flip rate and the coverage gap
are both read off pressured data, an association between them could be definitional. It
is not: Table~\ref{tab:screentransfer} measures the flip rate on one task type and
ranks pressured coverage on a different one. The off-diagonal entries have median
$\rho=-0.49$ with $35$ of $42$ in the right direction (sign test $p=7.5\times10^{-6}$),
matching the diagonal median of $-0.60$. Geometric shapes is the single exception: as a
prediction target its column is positive throughout, and dropping it leaves $35$ of $36$
correct.

\begin{table}[h]
\centering
\caption{Cross-task transfer. Row: the task the flip rate is measured on. Column: the
task whose pressured coverage is ranked. Off-diagonal entries are out-of-sample.
Seeds~1--2, six models with the complete task grid.}
\label{tab:screentransfer}
\small
\setlength{\tabcolsep}{4.5pt}
\begin{tabular}{lccccccc}
\toprule
Measured on & ARC & TQA & MMLU-Pro & LogDed & Track & Temporal & Geom \\
\midrule
ARC       & $-0.94$ & $-0.49$ & $-0.54$ & $-0.71$ & $-0.31$ & $-0.94$ & $+0.09$ \\
TQA       & $-0.83$ & $-0.20$ & $-0.43$ & $-0.77$ & $-0.43$ & $-0.83$ & $+0.26$ \\
MMLU-Pro  & $-1.00$ & $-0.66$ & $-0.60$ & $-0.77$ & $-0.37$ & $-1.00$ & $+0.03$ \\
LogDed    & $-0.83$ & $-0.60$ & $-0.31$ & $-0.77$ & $-0.49$ & $-0.83$ & $+0.31$ \\
Track     & $-0.67$ & $-0.49$ & $-0.14$ & $-0.23$ & $+0.14$ & $-0.67$ & $+0.26$ \\
Temporal  & $-0.71$ & $-0.94$ & $-0.09$ & $-0.26$ & $-0.03$ & $-0.71$ & $+0.20$ \\
Geom      & $-0.49$ & $-0.31$ & $+0.26$ & $-0.26$ & $-0.03$ & $-0.49$ & $+0.71$ \\
\bottomrule
\end{tabular}
\end{table}

What the two tables support is a ranking, not a cutoff. With six models per regime, a
binary split that happens to sort coverage arises by chance about a third of the time,
so we do not propose a flip-rate threshold below which a model is safe, and four
families is too few to read even the ranking as settled. The usable claim is narrower
and is the one a deployer needs when choosing between models: pressure has to be applied
to see vulnerability, and the task it is applied on need not be the deployment task.

\section{Qwen size scaling: model size is not a safe proxy}
\label{app:qwensize}

Seeds~1 and~2 include three Qwen variants (1.5B, 3B, 7B). A natural hypothesis is
that larger models are more robust, but the data contradict it
(Table~\ref{tab:qwensize}). Qwen-1.5B and Qwen-3B have near-identical answer-flip
rates ($\approx32\%$) yet coverage outcomes differing by $51$\,pp ($96.0\%$ vs.\
$45.2\%$), likely reflecting differences in the shape of the round-1 confidence
distribution (Qwen-3B is sharper, setting a tighter global threshold that amplifies
the pressure-induced drop). Model size is not a safe proxy for conformity
susceptibility. The $51$\,pp spread is itself mostly a shared-threshold artifact: it shrinks to
$8$\,pp once each size is calibrated on its own scores, so the two sizes differ far less
in conformity than the table suggests.

\begin{table}[h]
\centering
\caption{Qwen size scaling (seeds~1--2, all datasets, $\alpha=0.10$). The coverage
column uses the threshold shared by the three sizes, which is what makes the spread
below as wide as it is; on per-size thresholds it narrows to $8$\,pp.}
\label{tab:qwensize}
\small
\begin{tabular}{lcccc}
\toprule
Model & Flip rate & $\bar{\Delta p}_{\mathrm{gt}}$ & Coverage (UW) & Gap \\
\midrule
Qwen-1.5B & 32.8\% & $-$0.301 & \textcolor{ok}{96.0\%} & $+$6.0pp \\
Qwen-3B   & 31.3\% & $-$0.290 & \textcolor{collapse}{45.2\%} & $-$44.8pp \\
Qwen-7B   & 53.7\% & $-$0.528 & \textcolor{collapse}{48.1\%} & $-$41.9pp \\
\bottomrule
\end{tabular}
\end{table}

\section{Canonical numbers and reproducibility}
\label{app:numbers}

All numbers below are from CPU-only analysis of pre-computed LLM outputs. Every number
in the paper is regenerated from the released analysis code (\url{https://github.com/yibo-hu-lab/conformity-breaks-conformal}), and the headline
values are
cross-checked by a second implementation.

\begin{center}
\begin{tabular}{ll}
\toprule
Fact & Value \\
\midrule
Marginal coverage at round 1 (target 90\%) & $90.0\% \pm 0.0\%$ \\
Marginal coverage at unanimous-wrong (target 90\%) & $\mathbf{73.9\% \pm 0.3\%}$ \\
Gap & $-16.1$\,pp \\
Same gap, broader data (6 models, 7 task types) & $-20.2$\,pp \\
Marginal coverage at round 1 (target 95\%) & $95.0\% \pm 0.0\%$ \\
Marginal coverage at unanimous-wrong (target 95\%) & $86.4\% \pm 0.8\%$ \\
Gap & $-8.6$\,pp \\
No collapse at $\alpha=0.01$ (target 99\%) & $\approx 100\%$ all conditions \\
\midrule
Targeted band, no pressure (pooled) & $87.3\%$ (target 90\%) \\
Targeted band, unanimous-wrong (pooled) & $\mathbf{47.4\%}$ (target 90\%) \\
Same band, adversary probing the system & $79.3\%\to53.5\%$ \\
Bottom-20\% (reference, straddles threshold) & $35.5\%$ \\
Bottom-10\% (reference, below threshold) & $23.6\%$ \\
Singleton-ACT on attacker's answer (pooled) & $12.1\%$ of flipped items \\
Worst single model (Qwen-7B, seed-0, shared threshold) & $33.9\%$ (target 90\%) \\
Same model, three seeds pooled & $37.5\%$ \\
Same model, its own threshold & $61.0\%$ \\
Seeds used (marginal results) & 3 \\
\bottomrule
\end{tabular}
\end{center}

\section{Proof of Proposition~\ref{prop:coverage}}
\label{app:proof}

\paragraph{Part (i).}
Couple the clean and pressured scores on the same items and
decompose $\Delta=\Pr[s_0\le\hat\tau]-\Pr[s_1\le\hat\tau]$ by whether each item
crosses $\hat\tau$: covered items that fall out (down-crossings) raise $\Delta$,
uncovered items pulled in (up-crossings) lower it, and items that do not cross
cancel. Hence $\Delta$ is the net (down minus up) crossing flow.

\paragraph{Part (ii).}
$\hat\tau$ is the $(1-\alpha)$-quantile of $\mathcal{D}_0$, so $F_0(\hat\tau)=1-\alpha$
by definition (exactly so under continuity at $\hat\tau$; the empirical LLM scores
have atoms, e.g.\ at $p_{\mathrm{gt}}=1$, so finite-sample coverage carries the usual
$\pm 1/(n{+}1)$ slack). The hypothesis $F_1(\hat\tau)\le F_0(\hat\tau)$ then gives
$F_1(\hat\tau)\le 1-\alpha$ directly; stochastic dominance implies it but is not
needed. \hfill$\square$

\end{document}